\PassOptionsToPackage{numbers,sort&compress}{natbib}
\documentclass{article}

\usepackage[main, final]{neurips_2026}

\usepackage[utf8]{inputenc} %
\usepackage[T1]{fontenc}    %
\usepackage{hyperref}       %
\usepackage{url}            %
\usepackage{booktabs}       %
\usepackage{amsfonts}       %
\usepackage{nicefrac}       %
\usepackage{microtype}      %
\usepackage{xcolor}         %

\usepackage{algorithm}
\usepackage[noend]{algorithmic}

\usepackage{amsmath}
\usepackage{amssymb}
\usepackage{mathtools}
\usepackage{amsthm}

\usepackage[capitalize,noabbrev]{cleveref}

\theoremstyle{plain}
\newtheorem{theorem}{Theorem}[section]
\newtheorem{proposition}[theorem]{Proposition}
\newtheorem{lemma}[theorem]{Lemma}

\theoremstyle{definition}

\theoremstyle{remark}

\usepackage[textsize=tiny]{todonotes}
\usepackage[inline]{enumitem}
\usepackage{subcaption}
\usepackage{thmtools} 
\usepackage{thm-restate}

\usepackage{xcolor}

\newcommand{\eps}{\varepsilon}

\newcommand{\R}{\ensuremath{\mathbb{R}}}

\newcommand{\Var}{\ensuremath{\textsf{Var}}}
\newcommand{\Cov}{\ensuremath{\textsf{Cov}}}

\newcommand{\norm}[1]{\lVert #1 \rVert}

\newcommand\numberthis{\addtocounter{equation}{1}\tag{\theequation}}
\newcommand{\abs}[1]{| #1 |}

\newcommand{\inner}[2]{\langle #1,#2 \rangle}

\newcommand{\E}{\ensuremath{\textsf{E}}}
\newcommand{\G}{G} %

\newcommand{\Tr}{\ensuremath{\textsf{Tr}}}

\usepackage{bm}

\DeclareMathOperator{\CV}{CV}
\DeclareMathOperator{\ACV}{ACV}

\renewcommand{\ne}{\mathsf{ne}}
\newcommand{\wtheta}{\widehat{\theta}}
\newcommand{\wTheta}{\widehat{\Theta}}
\newcommand{\wG}{\widehat{\G}}

\newcommand{\bX}{\mathbf{X}}

\newcommand{\wne}{\widehat{\ne}}
\newcommand{\sign}{\mathsf{sign}}

\newcommand{\svmin}{\sigma_{\min}}
\newcommand{\svmax}{\sigma_{\max}}

\newcommand{\lambdao}{\lambda_*}
\newcommand{\lambdacv}{\lambda_{\CV}}
\newcommand{\lambdaacv}{\lambda_{\ACV}}

\newcommand{\bZ}{\mathbf{Z}}

\newcommand{\argmin}{\arg\min}
\DeclareMathOperator{\supp}{supp}

\DeclareMathOperator{\subG}{subG}

\title{Fast and Consistent Structure Learning in Graphical Models via Approximate Cross-Validation}

\author{%
  Zhao Lyu \\
  University of Chicago\\
  \texttt{zlyu@uchicago.edu} \\
  \And
  Mladen Kolar \\
  USC \& MBZUAI\\
  \texttt{mkolar@marshall.usc.edu} \\
  \And
  Bryon Aragam \\
  University of Chicago \\
  \texttt{bryon@chicagobooth.edu} \\
}

\begin{document}

\maketitle

\begin{abstract}
  Structure learning for sparse graphical models in high dimensions requires both computational scalability and selection consistency. Classical criteria such as (Extended) Bayesian Information Criterion can provide consistency, but rely on model-specific likelihoods and complexity calibration. In comparison, cross-validation avoids such analytic calibration and is therefore attractive for graph learning. However, cross-validation is computationally infeasible and theoretically inconsistent in structure learning. This creates a tension between consistency and flexibility in modeling. In this work, we resolve the tension by developing a scalable graph-recovery procedure built on approximate cross-validation. We demonstrate its effectiveness in learning large graphical models both theoretically and empirically, and establish selection consistency under mild signal strength assumptions.
\end{abstract}

\section{Introduction}

High-dimensional structure learning aims to recover latent structure, not merely to optimize predictive metrics. In graphical models, this structure is encoded by an edge set. The central challenge is therefore a selection problem: Among many candidate structures, we need a rule that is both computationally feasible and statistically correct.

Classical likelihood-calibrated criteria such as the Bayesian Information Criterion and Extended Bayesian Information Criterion (BIC~\cite{schwarz1978estimating}, EBIC~\cite{foygel2010extended}) can provide consistency guarantees, but they require a model-specific likelihood and complexity calibration in the form of a penalty. Cross-validation (CV) offers a different route~\citep[][]{stone1978cross, arlot2010survey, lei2020cross, bates2024cross}: It is model-agnostic in the sense that, once a validation loss is specified, the same empirical rule can compare candidate fitting procedures without deriving a bespoke penalty. It is also empirically calibrated: It evaluates candidates on held-out sample, rather than relying on a fixed analytic penalty derived for a specific model class. These properties make CV appealing for structure learning, but exact CV requires repeated refitting which is computationally infeasible for large graphical models.

Approximate cross-validation (ACV) directly targets this computational bottleneck by replacing repeated refits with inexpensive approximations~\citep[][]{beirami2017optimal, wang2018approximate, koh2019accuracy, rad2020scalable}. Existing theory mainly asks whether these approximations preserve validation losses or model-assessment quality~\citep[][]{stephenson2020approximate,stephenson2020approximatelowrank,wilson2020approximate}. For structure learning, however, loss approximation is not enough: A candidate can have good held-out loss while at the same time selecting the wrong structure. Thus the key question is not whether ACV mimics CV in estimating held-out loss, but whether the true structure is recoverable from the structure it selects.

In this work, we provide a comprehensive characterization of ACV as a structure learning rule, and give structure-level answers to these questions. ACV alone is not a consistent structure learning rule since it inherits the over-selection behavior of CV. This failure, however, turns out to be useful rather than fatal. A less cynical interpretation of over-selection is that it still provides a high quality initial screening. We show that the ACV-selected structure contains the true structure with high probability, making ACV a scalable intermediate estimate for graph recovery. Naturally, by following this with a proper post-screening step, we can remove extra edges and achieve selection consistency. We summarize our main contributions as follows:
\begin{enumerate}
    \item \textbf{A negative result: Inconsistency.} We analyze why ACV alone fails as a final structure-learning rule: It inherits the selection inconsistency of CV through over-selection.
    \item \textbf{A positive result: Post-screening consistency.} We view the ACV's failure complementarily, and treat ACV as a scalable screening. A proper post-correction built on the intermediate ACV estimate removes extra edges and yields the selection consistency.
    \item \textbf{Empirical validation.} We empirically demonstrate the computational-statistical tradeoff: ACV provides a fast high-recall screen, while the corrected procedure improves final graph recovery.
\end{enumerate}

\label{sec:bg}
\subsection{Problem set-up}
\label{sec:setup}

Let $\subG(\sigma^2)$\footnote{A zero-mean random variable $Z$ is sub-Gaussian if there exists $\sigma\in(0,\infty)$ such that $\E[e^{tZ}]\leq e^{t^2\sigma^2/2}~\forall t\in\R$. Here $\subG(\sigma^2)$ consists of the zero-mean laws $P$ such that, for $X\sim P$ with covariance $\Sigma$, each $X_i/\sqrt{\Sigma_{ii}}$ satisfies the above sub-Gaussian bound with $\sigma$.} denote the family of zero-mean sub-Gaussian distributions on $\mathbb{R}^p$ with parameter $\sigma^2$. Fix $P\in\subG(\sigma^2)$ and let $X = \begin{bmatrix}
X_1 , X_2 , \dots , X_{p}
\end{bmatrix}\sim P$.
The conditional independence structure of $X$ is represented by a Markov random field, i.e. an undirected graph $G = (V,E)$ with node set $V=[p]$. Specifically, $X_i$ and $X_j$ are conditionally independent given the remaining variables $X_{\setminus \{i,j\}}$ if there is no edge between node $i$ and node $j$. This is the local Markov property: each variable is conditionally independent of all its non-neighbors given its graph neighbors.

We consider learning the structure of $G$ from $n$ i.i.d.~observations of $X$, denoted by the $n\times p$ data matrix $\bX$. 
Each row of the assembled data matrix $\bX$ is an i.i.d. sample drawn from $P$, and $\Sigma = \Cov(X) \succ 0$ denotes the population covariance. 
If $X\sim \mathcal{N}(\bm{0},\Sigma)$, the precision matrix $\Sigma^{-1}$ encodes the edge structure of a graph via its zero pattern: $\Theta_{ij} = 0$ if and only if $(i,j)\not\in E$. 
This defines a valid Markov random field $G$ for $X$, and recovering $G$ reduces to estimating the zero pattern of $\Sigma^{-1}$. Many methods such as the Cholesky factor-based method~\citep[][]{wu2003nonparametric,huang2006covariance,bickel2008covariance} and penalized likelihood methods~\citep[][]{yuan2007model,friedman2008sparse,d2008first,rothman2008sparse} are used to estimate the precision matrix for structure learning. In particular, for any matrix $\bX\in\mathbb{R}^{n \times p}$ and $\lambda>0$, the popular $\ell_1$ penalized negative log likelihood estimator~\citep[][]{banerjee2008model,friedman2008sparse,ravikumar2011high} is defined by:
\begin{align*}
    \wTheta^{\lambda,\bX} 
    :&= \argmin_{\Theta\succ 0} -\log\det\Theta + \Tr(S(\bX)\Theta) +\lambda\norm{\Theta}_1,\numberthis\label{eq:obj func}
\end{align*}
where $S(\bX) = \bX^\top\bX/n$ is the empirical covariance matrix depending on $\bX$\footnote{$S(Y) = YY^\top$ for a single column vector $Y$. We abuse the $i$-th row $\bX^{(i)}$ as a column vector when denoting $S(\bX^{(i)})$.}, and $\norm{\Theta}_1 = \sum_{i,j}\abs{\Theta_{ij}}$. The estimated graph (zero pattern) is defined by 
\begin{align*}
    \wG^{\lambda,\bX}
    :&= \supp(\wTheta^{\lambda,\bX}) = \{(i,j): \wTheta^{\lambda,\bX}_{ij}\neq 0\}\label{eq:wG}\numberthis.  
\end{align*}

It is worth noting that the exact precision-matrix interpretation is tied to the usual Gaussian graphical model setting and related semiparametric extensions. For non-Gaussian data, the zero pattern of $\Sigma^{-1}$ need not encode conditional independence without additional assumptions, although precision-based estimators are still widely used to recover structural relationships:
For example, copula-based methods~\citep[][]{liu2009nonparanormal}, rank-based methods~\citep[][]{xue2012regularized,fan2017high} and robust penalties~\citep[][]{loh2012structure}.

In this work, we study sparse undirected graphical model selection through the support of the precision matrix estimator in~\eqref{eq:obj func}. Sub-Gaussianity is used to simplify the concentration arguments; robustness experiments and stress tests in Section~\ref{sec:weak signal and real} and Appendix~\ref{sec:appendix lognormal} and \ref{sec:appendix ising} demonstrate the applicability of our approach beyond sub-Gaussian assumptions.

\subsection{Cross-validation for graph selection}

In sparse graphical models, selecting $\lambda$ is equivalent to selecting the graph $\wG^{\lambda,\bX}$ in~\eqref{eq:wG}. Thus CV directly determines the learned graph.
Specifically, leave-one-out CV constructs $n$ estimates of $\Theta$ by holding out a single sample each time, and evaluates the performance of the estimate on this held-out sample. We denote $\bX^{(-i)}$
(resp. $\bX^{(i)}$) as the $(n-1)$-by-$p$ submatrix of $\bX$ by excluding its $i$th row for training (resp. the $i$th row of $\bX$ for validation) for $i\in [n]$. The selected graph is $\wG^{\lambdacv^\bX,\bX}$, where the selected penalty is defined by:
\begin{align*}
    \lambdacv^\bX
    := \arg\min_{\lambda>0}
    \frac{1}{n}\sum_{i=1}^n -\log\det&\wTheta^{\lambda,\bX^{(-i)}} +  \Tr\left(S(\bX^{(i)})\wTheta^{\lambda,\bX^{(-i)}}\right).
\end{align*}

As a tool in the technical analysis, we also define the oracle penalty choice $\lambdao^\bX$:
\begin{align*}
    \lambdao^\bX
    :=
    \arg\min_{\lambda>0}
    \E_{Y\sim P}-\log\det\wTheta^{\lambda,\bX} + \Tr\left(S(Y)\wTheta^{\lambda,\bX}\right)\numberthis\label{eq:lambdao}.
\end{align*}

Although different splitting strategies lead to a balance between computational cost and accuracy, the core cost of CV is repeated refitting. Many existing works reduce this cost by constructing the leave-one-out estimate from a warm start. For example, a Newton's step from the full-data estimator is used to approximate the optimized objective based on a training set~\citep[][]{beirami2017optimal,rad2020scalable,wilson2020approximate,stephenson2020approximate,stephenson2020approximatelowrank}. These approximations provide a computational route to the CV-selected graph, but the statistical question remains: Does the resulting graph recover the correct structure?

We use ACV to address the computational barrier, and then analyze the graph selected by this approximation. We focus on LOO-CV because it evaluates the model performance accurately compared to other CV variants~\citep[][]{arlot2010survey, rad2020scalable}; a $K$-fold version is used in large-scale experiments.

The rest of the paper is organized as follows. We define the ACV-selected graph in Section~\ref{sec:method}. Section~\ref{sec:acv inconsistent} and ~\ref{sec:pre screen} show why ACV alone fails to be consistent, although it remains useful as a screened candidate graph. In Section~\ref{sec:fdr control} we propose a correction step that restores selection consistency, and Section~\ref{sec:exp} empirically evaluates the resulting computational-statistical tradeoff, and validates our procedure for controlled recovery.

\section{Approximate cross-validation for graph selection}
\label{sec:method}
\label{sec:acv consrtuction}
Let $\widehat{\mathcal{S}} := \supp(\wTheta^{\lambda,\bX})$ be the support of the full data estimator for $\lambda > 0$. We construct an approximation $\widetilde{\Theta}^{\lambda,\bX^{(-i)}}$ to the LOO-CV estimator $\wTheta^{\lambda,\bX^{(-i)}}$ by using the full data estimator as a warm start:
\begin{align*}
    \widetilde{\Theta}^{\lambda,\bX^{(-i)}}
    := 
    &\begin{bmatrix}
        \wTheta^{\lambda,\bX}_{\widehat{\mathcal{S}}} + \left(\wTheta^{\lambda,\bX}\otimes \wTheta^{\lambda,\bX}\right)_{\widehat{\mathcal{S}}\widehat{\mathcal{S}}}\left(S(\bX) - S(\bX^{(-i)})\right)_{\widehat{\mathcal{S}}}\\
        \mathbf{0}_{\setminus\widehat{\mathcal{S}}}
    \end{bmatrix}
    \numberthis\label{eq:acv},
\end{align*}
where $\otimes$ denotes the Kronecker product and coordinate
restrictions are interpreted under vectorization. We use the
full-data optimizer as a warm start and apply the inverse Hessian
of the smooth likelihood term between restrictions to
$\widehat{\mathcal S}$. The support restriction is motivated by
ACV constructions for nonsmooth penalties
\citep{stephenson2020approximate};
Appendix~\ref{sec: acv derivation} analyzes the displayed update
directly.

While the positive definiteness of $\wTheta^{\lambda,\bX}$ enables an explicit Hessian inverse calculation~\citep[][]{stephenson2020approximate}, matrix operations grow with the Kronecker-restricted support. In our implementation, we exploit sparsity by vectorizing the sample covariance difference rather than forming the full Kronecker product. All approximations to LOO-CV still scale with the number of held-out samples, so for large $n$ we use the analogous $K$-fold ACV score whose number of held-out fits is fixed by $K$~\citep[][]{meijer2013efficient}.

The ACV-tuned penalty $\lambdaacv^{\bX}$ is the one that minimizes the validation loss:
\begin{align*}
    \lambdaacv^{\bX}
    :=\argmin_{\lambda>0}\frac1n\sum_{i=1}^n\left[-\log\det\widetilde{\Theta}^{\lambda,\bX^{(-i)}}+
    \Tr\!\left(S(\bX^{(i)})\widetilde{\Theta}^{\lambda,\bX^{(-i)}}\right)\right].\numberthis\label{eq:lambdaacv}
\end{align*}
where the log-determinant is treated as invalid whenever its argument is not positive-definite by assigning value $+\infty$. If the ACV infimum is $-\infty$, we set $\lambdaacv^{\bX}=1+\max_j S(\bX)_{jj}$. In the definitions of $\lambdao^{\bX}$ and $\lambdaacv^{\bX}$, $\argmin$ denotes a measurable exact positive minimizer when attained, and otherwise a measurable positive point whose objective value is within $n^{-2}$ of the finite infimum. The existence and details are discussed in Appendix~\ref{sec:acv existence}. The ACV-selected graph is then $\wG^{\lambdaacv^\bX,\bX}$.

\section{Inconsistency of ACV}
\label{sec:acv inconsistent}
By construction, ACV aims to approximate CV estimates, and naturally inherits the structural inconsistency. To better understand the connection, we first show that the oracle penalty defined in \eqref{eq:lambdao} can be well estimated by the ACV-tuned $\lambda$ in \eqref{eq:lambdaacv}.

\begin{lemma}[Connection to the oracle]
\label{lem:acv}
For every $\delta>0$,
\begin{align*}
    \mathsf{Pr}_{\bX\sim P^{\otimes n} }(\abs{\lambdaacv^{\bX} 
    -
    \lambdao^{\bX}}<\delta)\to 1 \quad \text{as $n\to \infty$.}
\end{align*}
\end{lemma}
The limiting behavior of $\lambdaacv$ is useful when replacing CV with ACV for computational purposes. 
By bridging $\lambdaacv$ with $\lambdao$, we are able to evaluate ACV in structure learning through the lens of the oracle $\lambdao$. 

Consider the Gaussian model, where consistent graph recovery is possible under appropriate tuning of $\lambda$~\citep[][]{wainwright2009sharp, ravikumar2011high}. 

\begin{theorem}[ACV inconsistency]
\label{thm:inconsis_G}
    Assume that $p\ge2$ and $\Sigma\succ0$ are fixed and that each row of $\bX$ is an independent sample from $\mathcal{N}(0,\Sigma)$. If the associated graph $\G$ is not complete, then
\[
 \limsup_{n\to\infty}
 \mathbf{Pr}_{\bX\sim N(0,\Sigma)^{\otimes n}}
 \left(\widehat G^{\lambdaacv^{\bX},\bX}=G\right)<1.
\]
\end{theorem}

Theorem \ref{thm:inconsis_G} demonstrates that choosing $\lambda$ based on ACV is provably inconsistent for structure learning for the special Gaussian graphical model. It aligns with existing inconsistency of cross-validation selected tuning for structure learning, even for sufficiently sparse graphs \citep[][]{lyu2024inconsistency}. This is to be contrasted with existing results that demonstrate, e.g. risk consistency of ACV. 
In other words, the incorrect selection of $\lambda$ by ACV is what causes the inconsistency in structure learning, despite the existence of a properly tuned $\lambda$ that achieves consistency. 
This \emph{structure} \emph{in}consistency also exists alongside \emph{risk} consistency.
This leaves us to ask how and why this happens, and how it can be corrected.

\section{Pre-screening: An alternative perspective on inconsistency}
\label{sec:pre screen}
There are three cases to consider in order to explain selection inconsistency. On the two extremes, one may either miss true edges, or falsely include incorrect ones. The third case lies in between, missing true edges as well as including false edges. Based on Lemma ~\ref{lem:acv}, it is worth taking a look at the oracle. 

Consider $\Theta^{\lambdao, \bX}$ the oracle model that reaches the best model performance in terms of the risk. Missing any features would likely bring down its performance, contradictory to the definition~\eqref{eq:lambdao}. Alternatively, good predictive performance suggests that \emph{some} informative structure has been learned, and consistency might be recoverable. This intuition suggests that all the true edges must be recovered, ruling out the possibility of missing true edges. This provides a more informative perspective on the inconsistency.

\begin{theorem}[Pre-screening]
\label{thm:acv overselect}
    Assume $\Sigma\succ0$ and \ref{asump:signal strength}. Then
    $\wG^{\lambda_{\ACV}^\bX,\bX}$ defined in \eqref{eq:wG} has the pre-screening property: It selects every true edge with high probability, i.e.
    \begin{align*}
        \mathsf{Pr}_{\bX\sim P^{\otimes n}}(\G\subseteq \wG^{\lambda_{\ACV}^\bX,\bX}) \to 1\text{ as } n \to \infty.
    \end{align*}
\end{theorem}

Theorem~\ref{thm:acv overselect} follows from the population-risk bound established in
the proof of Lemma~\ref{lem:acv}; see Appendix~\ref{sec:acv overselect} for a complete proof. We recall that Theorem~\ref{thm:inconsis_G} establishes the inconsistency of the ACV-selected graph, while Theorem~\ref{thm:acv overselect} establishes the pre-screening property. These results together show that the remaining discrepancies are extra edges rather than omitted true edges. 
Instead of interpreting this as a negative result, we take an alternative perspective that turns the inconsistency failure into an opportunity: Consistency is recoverable if false edges are successfully filtered out.

\section{Correcting the ACV-selected graph}
\label{sec:fdr control}
The quantity of interest is selection consistency of the graph estimate, i.e., whether the estimated graph recovers the true graph structure. The pre-screening property in Theorem~\ref{thm:acv overselect} suggests a filtering scheme to exclude falsely included edges in $\wG^{\lambdaacv^\bX, \bX}$, and distinguishability generally relies on the signal strength. To make the model selection problem identifiable, we make the following general assumptions to specify the high-dimensional graph structure recovery setup.

\begin{enumerate}[label=(A\arabic*)]
    \item \emph{High-dimensionality.}
    There exists $\tau>0$ so that 
    the dimension $p$ grows with $n$ with $p\leq C_1n^{\tau}$ for some constant $C_1>0$.\label{asump:high dim}
    
    \item \emph{Nonsingularity.} \label{asump:nonsingular}
    The covariance matrix is positive definite, $\Cov(X)\succ 0$, with bounded condition number $\kappa(\Sigma)\leq C_2$ for some constant $C_2 > 0$.
    
    \item \emph{Signal strength.}\label{asump:signal strength}
    There exists some constant $C_3 > 0$ and $0\leq\gamma<\frac{1}{2}$ such that for every $(i,j)\in E$,
    \begin{align*}
            \abs{\rho_{ij\mid\G}} 
            \geq C_3 n^{-\gamma},
    \end{align*}
    where $\rho_{ij\mid G}=-\frac{(\Sigma^{-1})_{ij}}{\sqrt{(\Sigma^{-1})_{ii}(\Sigma^{-1})_{jj}}}$ is the full-order linear partial correlation.
    \item \emph{Linearity.} \label{asump:indep}
    Let $\ne(j)$ denote the neighbors of node $j$ in $G$. For every $j\in V$, the conditional mean of $X_j$ given its graph neighbors is linear:
    $X_j = \theta_j^\top X_{\ne(j)} + \eps_j$ with $\eps_j\perp X_{\ne(j)}$ and $\E[\eps_j\mid X_{\ne(j)}]=0$.
\end{enumerate}

We pause to note \ref{asump:high dim}-\ref{asump:signal strength} are standard regularity conditions for high-dimensional recovery. \ref{asump:indep} is a simplifying assumption that keeps the theoretical target within sparse precision-graph recovery; analyzing general nonlinear conditional-independence graphs would require different nonparametric assumptions and is beyond the scope of this paper. The robustness experiments in Section~\ref{sec:appendix ising} evaluate how the procedure behaves outside this formal setup.

Under these assumptions, we apply the post-screening scheme in Algorithm~\ref{alg: post cleaning}, and conclude the consistency of corrected ACV estimate.

\begin{theorem}
\label{thm:two_stage}
    Let $\check G^{\lambdaacv^{\bX},\bX}$ denote the output of Algorithm~\ref{alg: post cleaning} with input $\widehat G=\widehat G^{\lambdaacv^{\bX},\bX}$ and $\alpha=n^{-\varepsilon}$ for a fixed $\varepsilon>2\tau$. Then under Assumptions \ref{asump:high dim}-\ref{asump:indep}
    \begin{align*}
        \mathsf{Pr}_{\bX\sim P^{\otimes n}}(\check{\G}^{\lambdaacv^\bX}=\G)\to 1 \text{ as } n\to\infty.
    \end{align*}
\end{theorem}
Theorem~\ref{thm:two_stage} shows that the additional post-correction step obtains a consistent graph estimate with $\lambdaacv$. We highlight that the consistency of $\check{\G}^{\lambdaacv,\bX}$ is not contradictory to existing inconsistency results. A straightforward observation is that CV processes and selects the model in one single step, but $\check{\G}$ involves graph estimation in two steps. In other words, ACV prepares sufficient information (pre-screening) to be refined in the second stage (post-correction).

\begin{algorithm}[t]
\caption{Post Correcting for ACV-tuned $\lambdaacv$}\label{alg: post cleaning}
\begin{algorithmic}[1]
\REQUIRE An estimated graph $\wG= (V,\widehat{E})$ with $\G\subseteq\wG$, level $\alpha \in (0,1)$, data matrix $\mathbf{X} \in \mathbb{R}^{n \times p}$
\ENSURE Refined graph estimate $\check{G}^{\alpha, \mathbf{X}}$
\FOR{each edge $(i,j) \in \widehat{E}$}
  \STATE Find the union of neighbors in $\wG$:
  $$U(i,j) = \{k \in [p] \setminus \{i,j\}: (i,k) \in \widehat{E} \vee (j,k) \in \widehat{E}\}$$
  
  \STATE Compute residuals $e_i$ (resp. $e_j$) from ordinary least-squares regression of $\mathbf{X}_i$ (resp. $\mathbf{X}_j$) on $\mathbf{X}_{U(i,j)}$
  
  \STATE Construct the normalized partial correlation test statistic as follows
  $$
  \begin{aligned}
      &\widehat{T}_{ij} = \sqrt{\frac{n}{\frac{1}{n}\sum_{k=1}^n\left(e^{(k)}_i e^{(k)}_j - \Cov(e_i, e_j)\right)^2}}\Cov(e_i, e_j)
  \end{aligned}$$
\ENDFOR
\STATE Construct the refined edge set and graph estimate given by 
\begin{align*}
    \check{E}^{\alpha, \mathbf{X}} 
    &= \left\{(i,j)\in\widehat{E} : \abs{\widehat{T}_{ij}}\geq \Phi^{-1}\left(1-\frac{\alpha}{2p^2}\right)\right\},\\
    \check{\G}^{\alpha, \mathbf{X}}
    &= (V, \check{E}^{\alpha, \mathbf{X}}),
\end{align*}
where $\Phi$ denotes the cumulative distribution function of the standard normal distribution.

\end{algorithmic}
\end{algorithm}

Essentially Algorithm~\ref{alg: post cleaning} constructs a normalized partial correlation statistic $\widehat{T}$ to test whether or not the true partial correlation is zero. Here $\Cov(e_i,e_j)=n^{-1}\sum_{k=1}^n e_i^{(k)}e_j^{(k)}$. Least-squares residuals are defined by orthogonal projection, and we set $\widehat T_{ij}=0$ if its denominator is zero in practice (details can be found in Appendix~\ref{sec:Tstat}). The general idea of normalized statistics is similar to many thresholding methods~\citep{cai2016large,banerjee2008model,luo2014sure,ma2020variable}. 
Under the null, the normalized statistic admits a uniform normal-tail
approximation over the range used by the testing threshold. We prove the effectiveness of our algorithm in the following Theorem \ref{thm:fdr control}:

\begin{theorem}
\label{thm:fdr control}
    Assume that we initialize Algorithm \ref{alg: post cleaning} with $\wG$ such that $\wG\supseteq \G$.
    Under Assumptions \ref{asump:high dim}, \ref{asump:nonsingular} and~\ref{asump:indep}, for $\alpha\in(0,1)$
    \begin{align*}
        \mathsf{Pr}_{\bX\sim P^{\otimes n}}(\check{\G}^{\alpha,\bX}\not\subseteq\G)\leq \alpha
        \quad
        \text{as $n\to\infty$,}
    \end{align*}
    where $\check{G}^{\alpha,\bX}$ is constructed via the Algorithm \ref{alg: post cleaning}.
\end{theorem}

Notably the requirement places no barriers for the ACV and other CV variants due to the pre-screening property in Theorem \ref{thm:acv overselect}. When concerning large sparse graphs, directly applying the whole graph would place the computational burden on independence testing, wasting scalable and efficient ACV to reduce the estimate edge set size. We also pause to note that post-correction step is widely adaptive via other schemes, such as ranking and perturbation tests. The key point here is that structure information is indeed recoverable from ACV, which makes an efficient, adaptive and model-agnostic approach to extract a guaranteed graph estimate.

\section{Experiments}
\label{sec:exp}

For demonstration, we set all comparisons on the k-nearest neighbor graph with data generated from different distributions and $k=1$. The true graph is constructed by connecting each vertex with its $k$-nearest neighbors based on a randomly generated metric. Varying graph types concern different sparsity, and we defer the detailed graph constructions to Appendix \ref{Expt setup}. 

We consider four methods in total: ACV incorporated with the post-correction in Algorithm~\ref{alg: post cleaning}, $K$-fold cross validation (CV $K$-fold), BIC and EBIC. Unless stated otherwise, we use $K=3$ for both ACV and CV, and leave comparison with other CV variants to the Appendix. The first two represent the model-agnostic methods, while the latter two represent classical consistent methods. We also include comparisons with stability selection~\citep[][]{liu2010stability3}, TIGER~\cite[][]{liu2017tiger}, and Lockhart's covariance test~\citep[][]{lockhart2014significance} in Appendix~\ref{sec:appendix other methods}. 
To cover different perspectives under the sparsity regularized method, we include the following dataset setup: linear Gaussian, linear non-Gaussian, nonlinear, weak signals, and a real gene dataset. We explicitly demonstrate structure learning performance at pre-screening and post-correction stage respectively, visualizing a computational and statistical tradeoff. We defer additional evaluation of non-sub-Gaussian and non-linear models to Appendix~\ref{sec:appendix lognormal}-\ref{sec:appendix ising}.

To evaluate each method, we report average True Positive Rate (TPR), average False Discovery Rate (FDR), and the wall clock time in seconds.
The set of tuning parameters is pre-computed to ensure that the maximal $\lambda$ provides an empty graph estimate, and the minimal $\lambda$ provides a fully-connected graph estimate. 
We allow the number of observations $n$ and the number of observed variables $p$ to vary independently so one can be greatly larger than another, reflecting the high dimensional setting. We also set a wall time limit (36 hours). Exceeding the wall time limit or undefined FDR values are marked as missing values.

\subsection{ACV as a scalable screened graph estimate}
\label{sec: knn pre-screen setup}
\begin{figure}[ht]
    \begin{center}    
    \centerline{\includegraphics[width=\columnwidth]{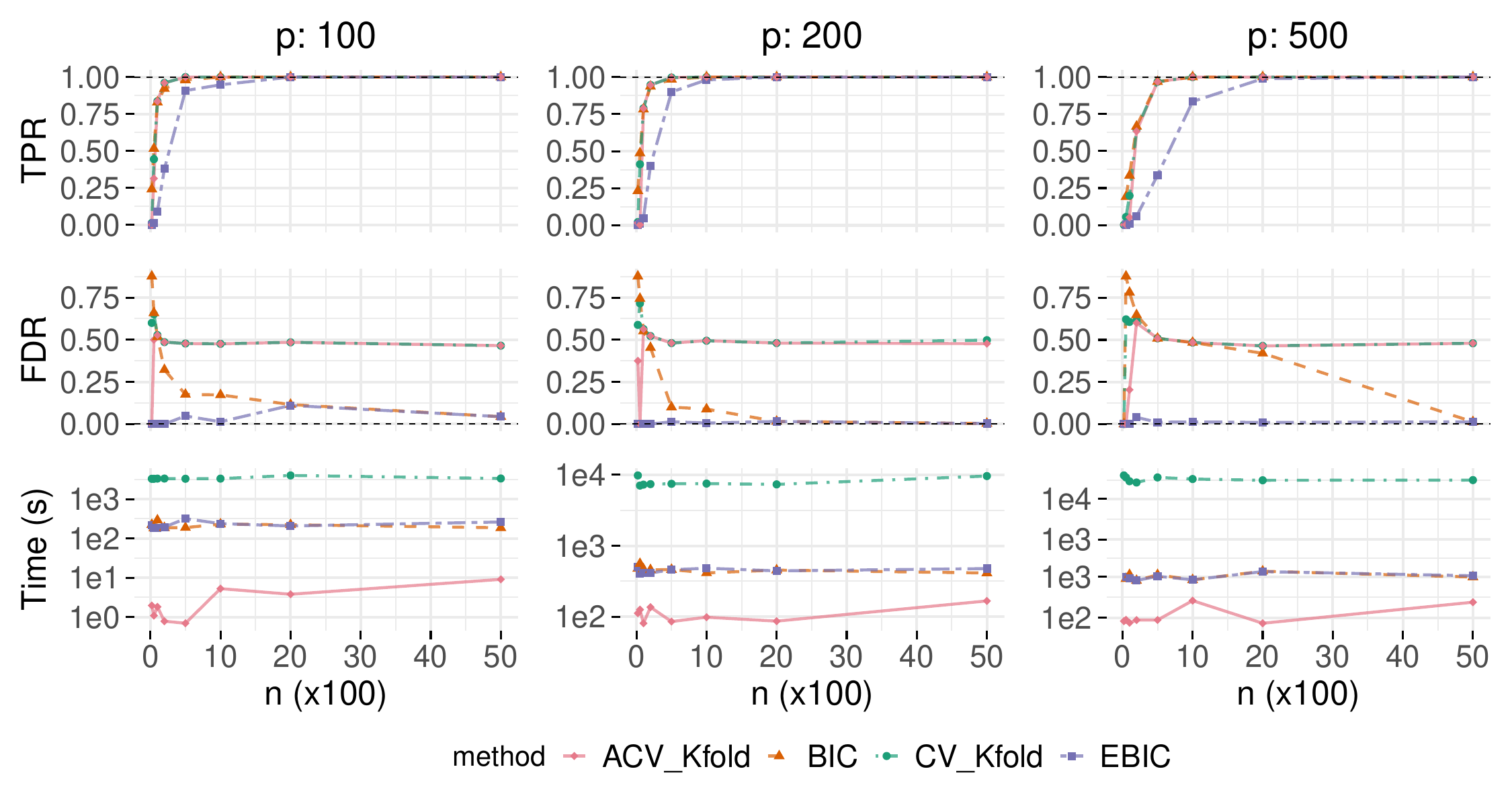}}
  \caption{Evaluation of ACV $K$-fold, CV $K$-fold and (extended) BIC in learning large scale KNN graph on sub-Gaussian data on pre-screening.}
  \label{fig: EXP2 lambdacv large scale}
  \end{center}
\end{figure}

We implement a $K$-fold version of ACV that averages over $K$ folds and compare it to the corresponding $K$-fold CV estimate.
We consider $p=(100,200,500)$ with sample size from $n=50$ to $n=5000$ to address a high dimensional setting and computational challenge.
We demonstrate structure learning performance between ACV,  for pre-screening and post-correction under Gaussian dataset, to separately present controlled consistency and computation advantage of ACV incorporated with the post-correction. These results are shown in Figure~\ref{fig: EXP2 lambdacv large scale}. 

\begin{figure}[ht]
    \begin{center}    \centerline{\includegraphics[width=\columnwidth]{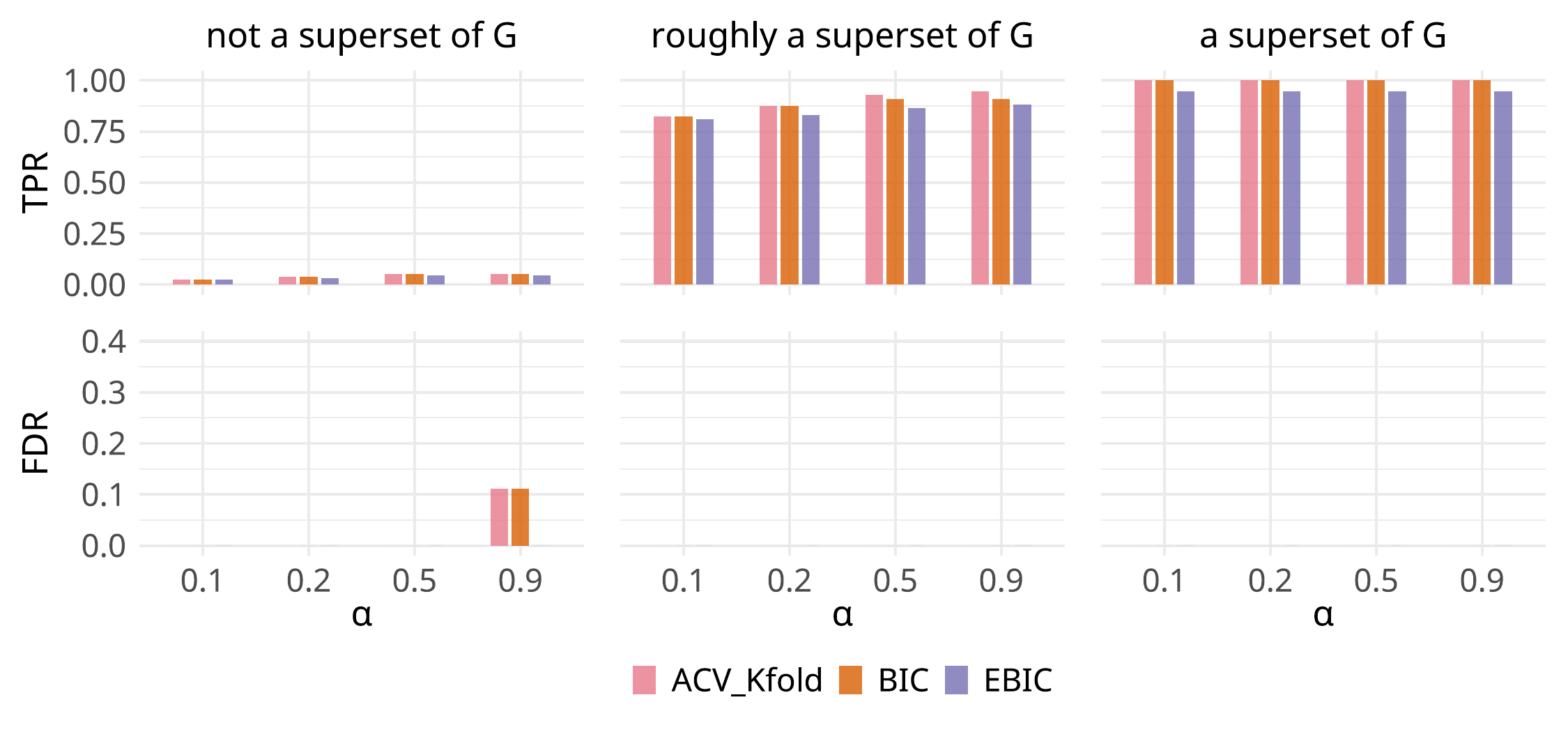}}
  \caption{TPR vs FDR after post-correction on estimates with $\lambda$ tuned by ACV $K$-fold, BIC and EBIC in large scale KNN graph on sub-Gaussian data.}
  \label{fig: EXP3 tpr fdr}
  \end{center}
\end{figure}

Figure~\ref{fig: EXP2 lambdacv large scale} shows that ACV $K$-fold closely tracks $K$-fold CV in TPR while taking much less time. This is the computational role of ACV: It gives access to a CV-like selected graph without repeated full refitting. The same figure also shows the statistical limitation of raw CV-type methods: Their FDR does not vanish once the TPR saturates. This matches Theorems~\ref{thm:inconsis_G} and~\ref{thm:acv overselect}: ACV is useful as a high-recall screened graph, but not as the final recovered graph. BIC and EBIC provide important likelihood-calibrated baselines; in these experiments they tend to enter true edges later along the sample-size axis, which motivates the correction of the ACV screen rather than replacing it with an information criterion.

\subsection{Correction of the ACV-selected graph}
\label{sec:exp_sub3}

We empirically evaluate the correction step by applying Algorithm~\ref{alg: post cleaning} to the screened graphs for $p=100$ in Figure~\ref{fig: EXP2 lambdacv large scale}. We use $\alpha \in\{0.1, 0.2, 0.5, 0.9\}$ and select $n=(100,500,1000)$ to represent three regimes for the input graph: (1) it misses true edges, (2) it approximately screens the true graph, and (3) it contains the true graph. This directly probes the sparse-screening requirement in Theorem~\ref{thm:fdr control}. The result is shown in Figure~\ref{fig: EXP3 tpr fdr}.

Figure~\ref{fig: EXP3 tpr fdr} shows the intended behavior of the correction. With limited samples at $n=100$, the screened graph misses true edges. No correction step can recover those omitted edges and TPR is reduced. With more samples at $n=500$, the screened graph approximately contains the true graph. In this case, the correction substantially reduces false discoveries while retaining most true edges. With sufficient sample size at $n=1000$, the screened graph contains the true graph, and the correction removes the extra edges and reaches zero FDR. This is the empirical counterpart of the theory: Screening quality is the first-order requirement, and correction addresses the over-selection of the raw CV-type graph.

\subsection{Weak signals and real data}
\label{sec:weak signal and real}
We next study settings where edge signals approach the boundary of the signal-strength condition. We use Gaussian data and KNN graphs with $k=1$. Following the signal strength parameter $\gamma$ in Assumption~\ref{asump:signal strength}, we fix $p=100$ and set the minimal partial correlation to scale like $n^{-\gamma}$ with $\gamma\in\{0,0.25,0.5\}$, where larger $\gamma$ corresponds to weaker signals.

\begin{figure}[ht]
    \begin{center}
    \centerline{\includegraphics[width=\columnwidth]{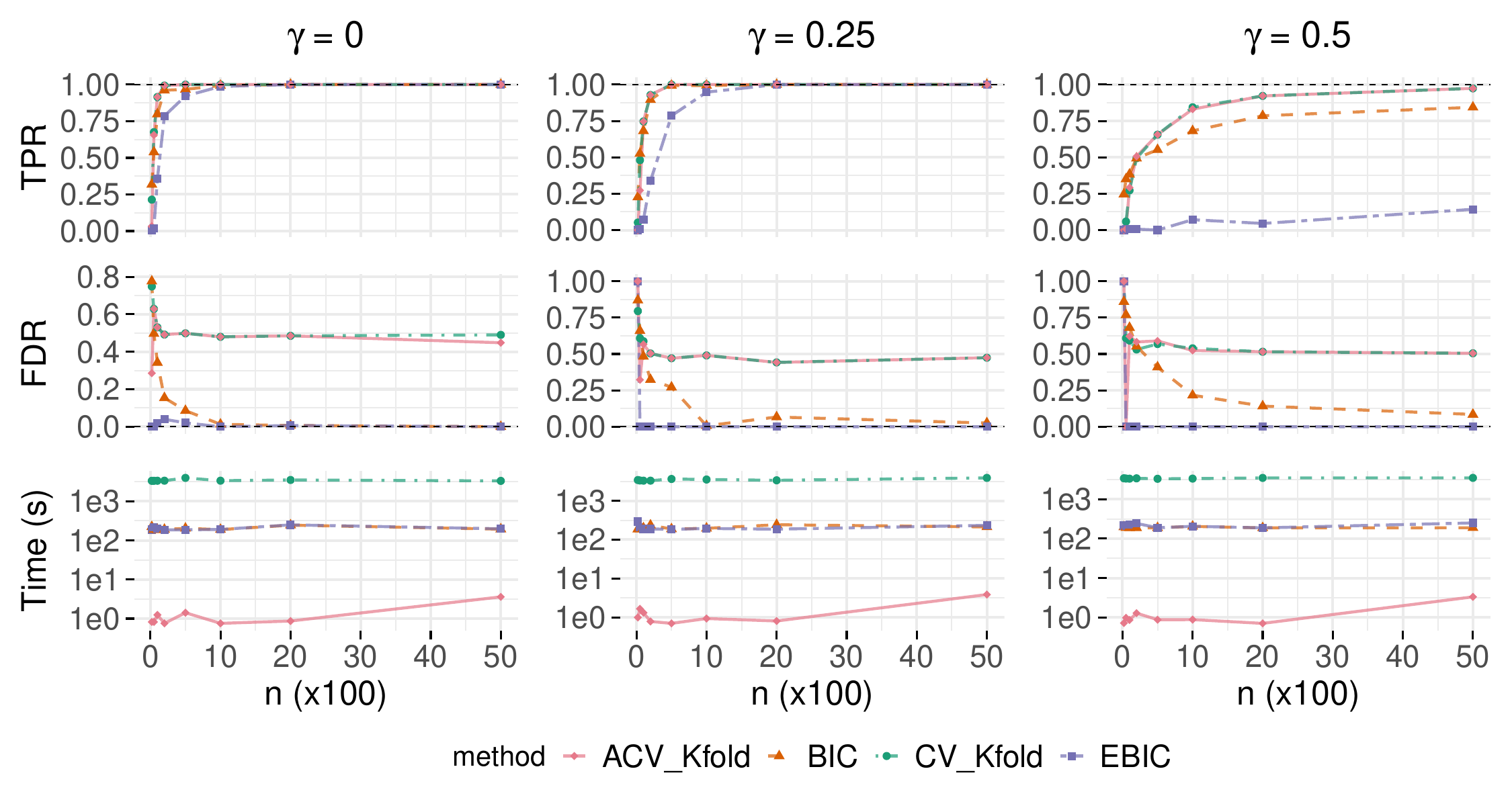}}
  \caption{Comparison of ACV $K$-fold, CV $K$-fold and (extended) BIC on large KNN graphs over varying signal strength $\gamma$.}
  \label{fig: EXP2 signal large scale}
  \end{center}
\end{figure}

Figure~\ref{fig: EXP2 signal large scale} shows that for strong or moderate signals ($\gamma=0,0.25$), all methods have near-perfect TPR and very small FDR, while ACV is much faster than $K$-fold CV. For weak signals ($\gamma=0.5$), EBIC detects less than one-third of the true edges and BIC still misses true edges at $n=5000$. In contrast, $K$-fold CV and ACV retain close to 90\% of true edges, with ACV requiring far less time than $K$-fold CV. This supports the role of ACV as the candidate-graph stage: Correction can only succeed when the first-stage graph retains the true edges.

\begin{table}[t]
  \caption{Gene-expression screened-graph comparison.}
  \label{tab:real_table}
  \begin{center}
    \begin{small}
      \begin{sc}
        \begin{tabular}{lccccr}
          \toprule
          Method & TPR & F1 & Time (s) \\
          \midrule
          BIC & 0.167   & 0.095 & 0.062 \\
          EBIC & 0.167   & 0.125 & 0.071 \\
          ProxACV & \textbf{1.000}   & \textbf{0.240} & 56.544 \\
          ACV & \textbf{1.000}   &\textbf{0.240} & 0.685 \\
          ACV $K$-fold & \textbf{1.000}   & \textbf{0.240} & \textbf{0.058} \\
          \bottomrule
        \end{tabular}
      \end{sc}
    \end{small}
  \end{center}
  \vskip -0.1in
\end{table}

\begin{figure}[ht]
  \centering
  \begin{subfigure}[t]{0.3\textwidth}
    \centering
    \includegraphics[width=\linewidth]{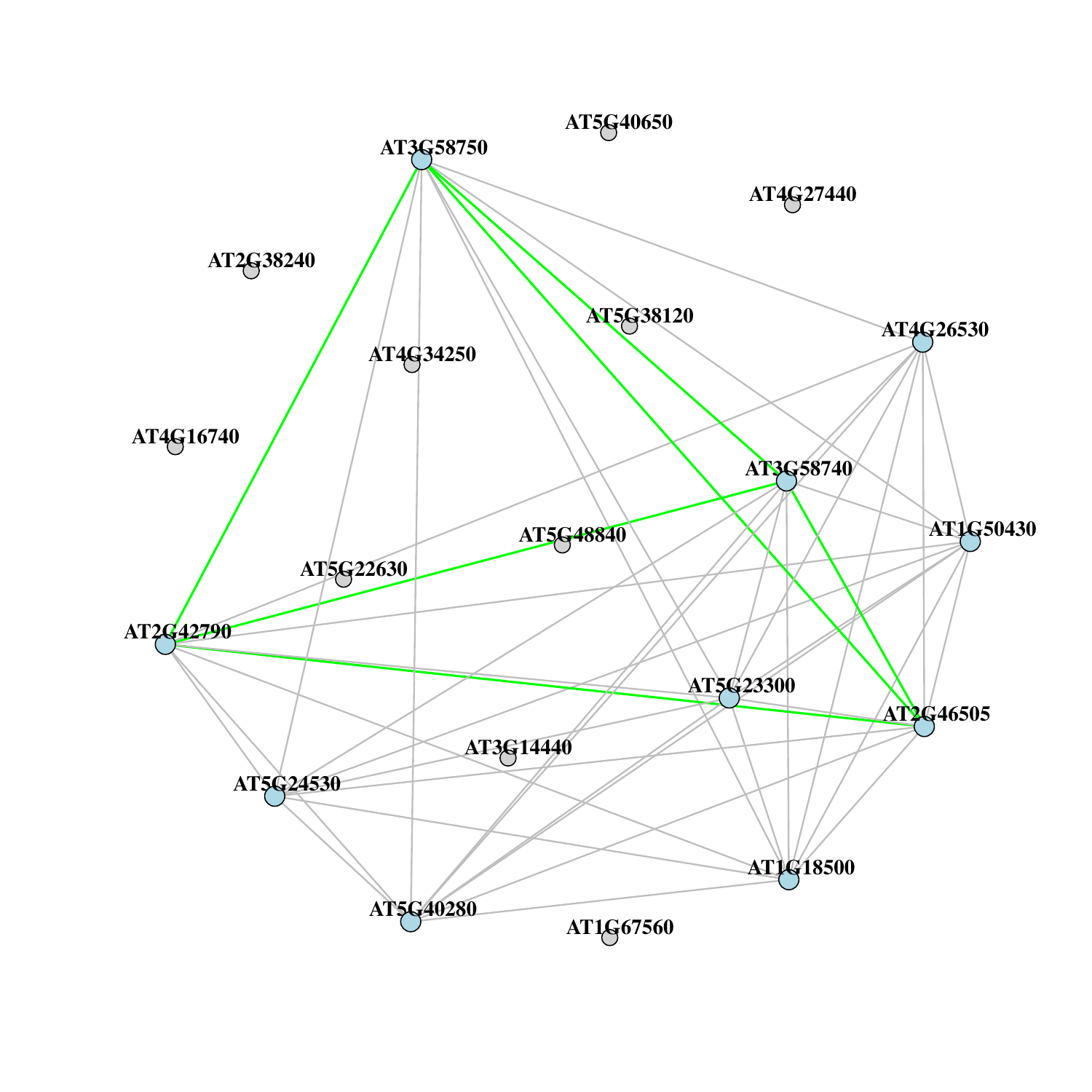}
    \caption{ACV $K$-fold}
  \end{subfigure}\hfill
  \begin{subfigure}[t]{0.3\textwidth}
    \centering
    \includegraphics[width=\linewidth]{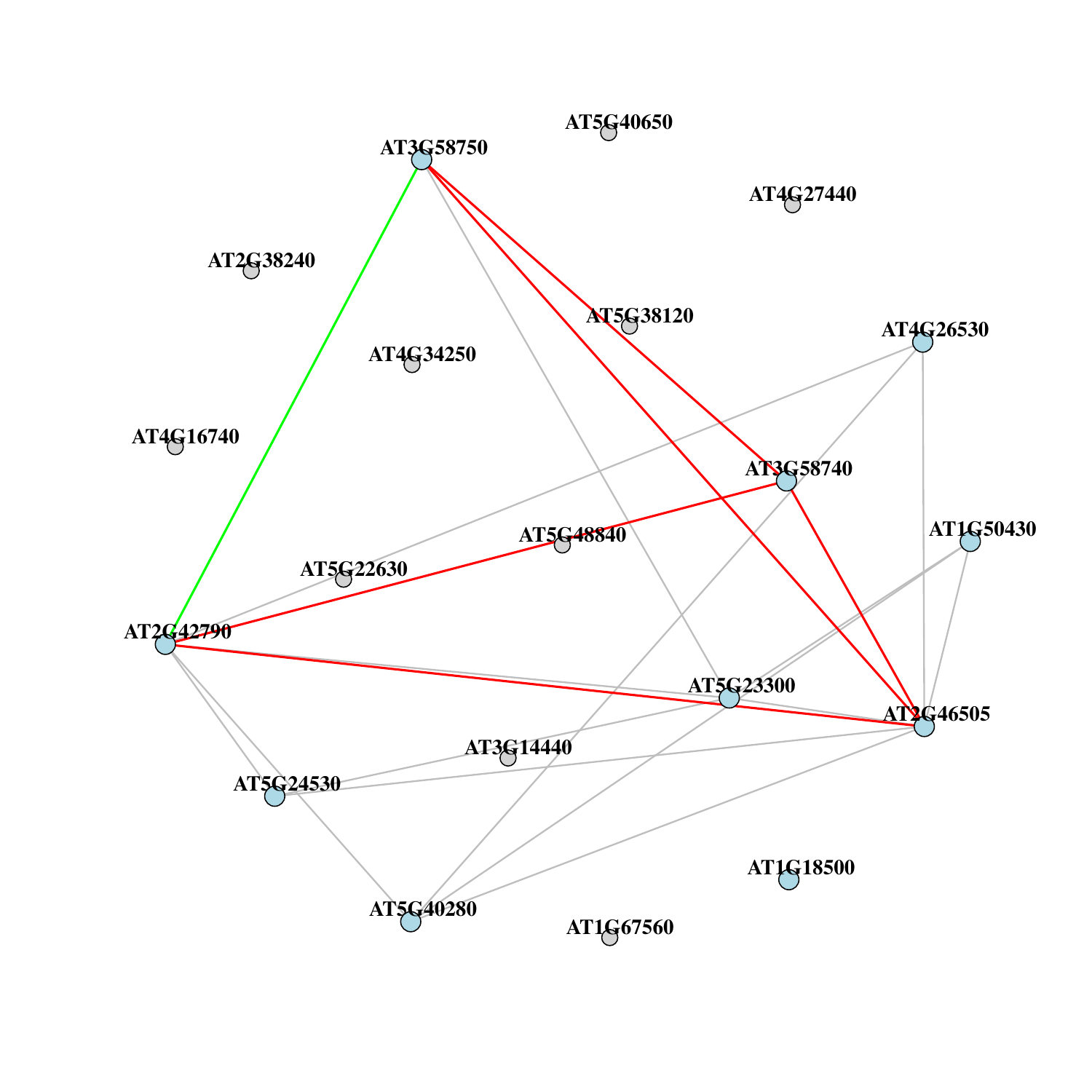}
    \caption{BIC}
  \end{subfigure}\hfill
  \begin{subfigure}[t]{0.3\textwidth}
    \centering
    \includegraphics[width=\linewidth]{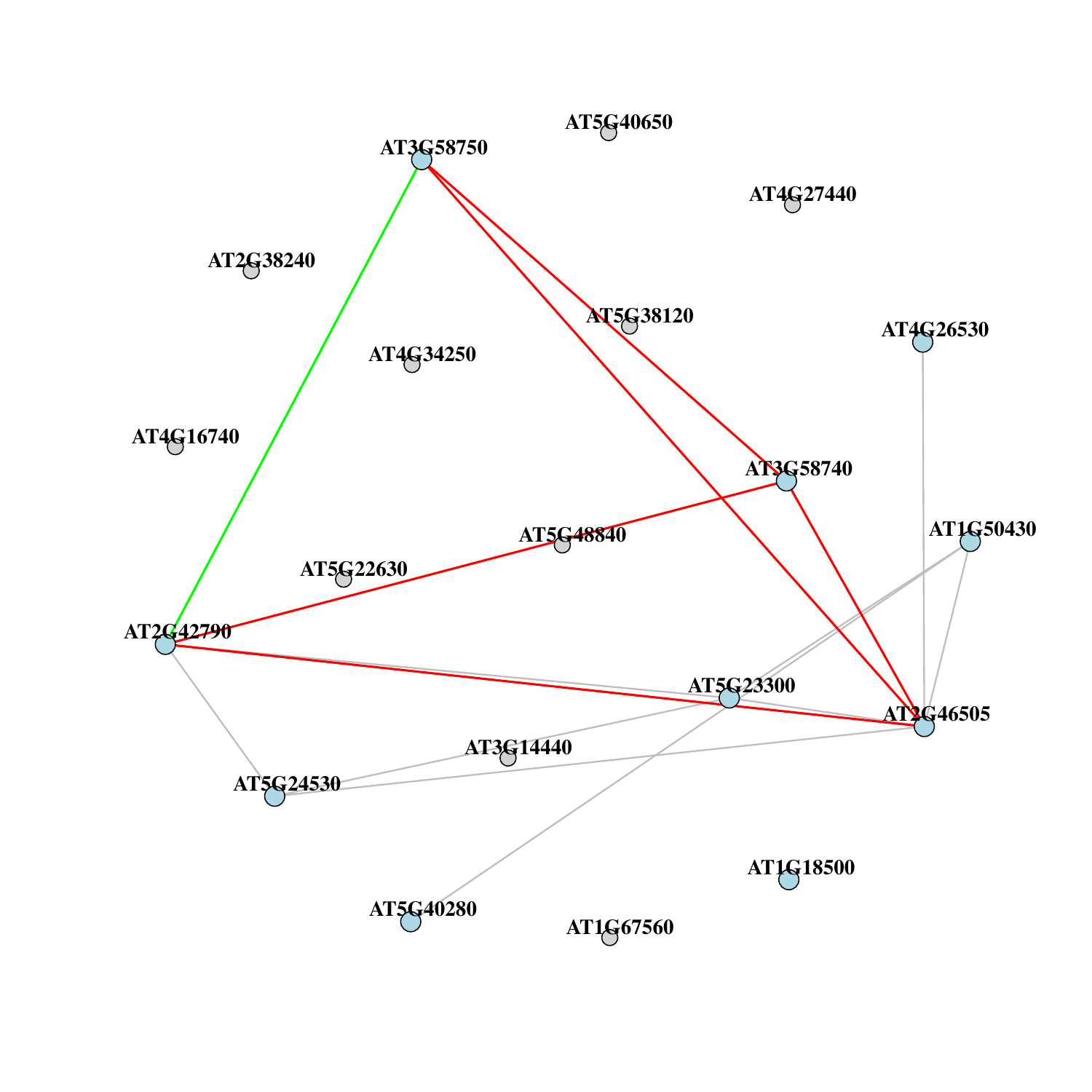}
    \caption{EBIC}
  \end{subfigure}
  \caption{Screened graphs selected by ACV $K$-fold, BIC, and EBIC on the reduced gene-expression graph. \textcolor{green}{Correctly identified edges} are marked in green and \textcolor{red}{missing edges} are marked in red.}
  \label{fig:real_graph}
\end{figure}

Finally, we apply the methods to an \emph{Arabidopsis thaliana} gene-expression dataset with known functional associations as a reference graph~\citep[][]{wille2004sparse,li2015gene,lee2010rational}. Table~\ref{tab:real_table} and Figure~\ref{fig:real_graph} report the screened-graph task. ACV and its variants retain all known edges, while BIC and EBIC miss more than 80\% of known edges. Appendices~\ref{sec:appendix lognormal}-\ref{sec:appendix ising} report stress tests under log-normal data and a nonlinear Ising model, and Appendix~\ref{sec:appendix other methods} compares with stability selection, TIGER, and Lockhart's covariance test.

\section{Conclusion and limitations}
\label{sec:conclusion}
In this paper, we studied the problem of graphical model selection in high-dimensions, where both statistical and computational issues arise. Specifically, we consider the use of cross-validation type procedures, which are inconsistent but have an appealing pre-screening property that can be exploited as part of a two-stage screen-and-correct strategy.
This allows fast approximations to cross-validation, such as approximate CV, to be used. We interpret the intrinsic inconsistency of ACV as a scalable intermediate estimate encoding useful structural information. The pre-screening property and inconsistency of ACV demonstrate the necessity of the post-correction procedure for model selection. Our theoretical results establish that, when incorporated with the post-correction step, ACV is in fact consistent in model selection, addressing both inconsistency and computational concerns inherent in traditional cross validation methods. This provides a desirable balance of correctness and efficiency in practice.

We propose one such approach in Algorithm~\ref{alg: post cleaning}, but there exists a wide range of post-correction strategies to build on the intermediate estimate constructed by ACV. Our current choice relies on partial independence testing and leans on a more conservative false discovery control. Alternative correction schemes may provide better trade-offs; adapting to other schemes and comparing for the best tradeoff is an interesting future direction.

Overall, empirical results validate the effectiveness of the chosen post-correction on ACV, showing substantial improvements in both consistency and computational efficiency compared to existing structure learning methods. The proposed method proves particularly advantageous in large scale graphical models, where traditional methods often struggle with scalability. Our work reveals fundamental structure information hidden within inconsistent procedures such as ACV, and naturally extend to an efficient and scalable framework for structure learning that balances theoretical rigor with practical performance, making it a valuable tool for high-dimensional graphical model selection.

\paragraph*{Acknowledgements}
BA was supported by NSF DMS-2610618, IIS-2453378, IIS-1956330, and the Robert H. Topel Faculty Research Fund.

\bibliography{ref}
\bibliographystyle{abbrvnat}

\appendix
\section{Analysis of the ACV estimate}
\label{sec:app A}
Let $P\in\subG(\sigma^2)$ be the zero-mean distribution specified
in Section~\ref{sec:setup}, and let $\bX\sim P^{\otimes n}$ be the
$n\times p$ data matrix, with $n\ge2$. Write $\Sigma=\Cov(X)\succ0$.
Let $\bX^{(i)}$ be the $i$th observation, interpreted as a column vector, and $\bX^{(-i)}$ omits that observation. The finite-sample estimates below retain their dependence on
$p$ and $\Sigma$. When $p$ and $P$ vary with $n$, all asymptotic conclusions
throughout the paper use the sufficient-sample regime in which
the explicitly displayed failure probabilities tend to zero. The standardized
marginal sub-Gaussian bound is common across the sequence.

\subsection{Derivation of $\widetilde{\Theta}^{\lambda,\bX}$}
\label{sec: acv derivation}

For each $i\in[n]$, we approximate $\wTheta^{\lambda,\bX^{(-i)}}$
using the full-data estimate. Under the population mean-zero
assumption in Section~\ref{sec:setup}, define
\begin{align*}
 F(\Theta,\lambda)
 &:= -\log\det\Theta+\Tr(S(\bX)\Theta)+\lambda\norm{\Theta}_1,\\
 F^{(-i)}(\Theta,\lambda)
 &:= -\log\det\Theta+\Tr(S(\bX^{(-i)})\Theta)
       +\lambda\norm{\Theta}_1,
\end{align*}
where
\[
 S(\bX)=\frac1n\sum_{i=1}^n\bX^{(i)}(\bX^{(i)})^\top,
 \qquad
 S(\bX^{(-i)})=\frac1{n-1}\sum_{j\neq i}
                  \bX^{(j)}(\bX^{(j)})^\top.
\]
Consequently,
\[
 \wTheta^{\lambda,\bX^{(-i)}}=
 \argmin_{\Theta\succ0}
 \left\{F(\Theta,\lambda)
 -\Tr\big((S(\bX)-S(\bX^{(-i)}))\Theta\big)\right\}.
\]

Let $\widehat{\mathcal S}=\supp(\wTheta^{\lambda,\bX})$.
The inverse Hessian of the smooth likelihood term at
$\wTheta^{\lambda,\bX}$ is
$\wTheta^{\lambda,\bX}\otimes\wTheta^{\lambda,\bX}$.
We apply this inverse Hessian between the coordinate restrictions
to $\widehat{\mathcal S}$ and define
\begin{align*}
 \widetilde{\Theta}^{\lambda,\bX^{(-i)}}_{\widehat{\mathcal S}}
 &:=
 \wTheta^{\lambda,\bX}_{\widehat{\mathcal S}}
 +(\wTheta^{\lambda,\bX}\otimes\wTheta^{\lambda,\bX})
       _{\widehat{\mathcal S}\widehat{\mathcal S}}
       (S(\bX)-S(\bX^{(-i)}))_{\widehat{\mathcal S}},\\
 \widetilde{\Theta}^{\lambda,\bX^{(-i)}}_
                  {\widehat{\mathcal S}^{\,c}}&:=0.
\end{align*}
Here the coordinate restrictions are interpreted under vectorization,
as in the main-text construction.
This is a support-masked inverse-Hessian update. 
The following analysis applies to
the displayed construction and does not assume equality of the
full-data and leave-one-out supports.

\subsection{Evaluation of ACV}
\label{sec:acv_cv_lossbound}

We compare the validation losses of ACV and exact leave-one-out CV.
Write $S_{-i}=S(\bX^{(-i)})$, $S=S(\bX)$, and $S_i=S(\bX^{(i)})$.
Define
\begin{align*}
 \ACV(\lambda)
 &:=\frac1n\sum_{i=1}^n\left[
 -\log\det\widetilde{\Theta}^{\lambda,\bX^{(-i)}}
 +\Tr(S_i\widetilde{\Theta}^{\lambda,\bX^{(-i)}})\right],\\
 \CV(\lambda)
 &:=\frac1n\sum_{i=1}^n\left[
 -\log\det\wTheta^{\lambda,\bX^{(-i)}}
 +\Tr(S_i\wTheta^{\lambda,\bX^{(-i)}})\right].
\end{align*}
The ACV score is $+\infty$ if any approximate fold is not positive
definite. Throughout this appendix,
\[
 \norm{A}_1=\sum_{j,k}|A_{jk}|,
 \qquad \norm{A}_\infty=\max_{j,k}|A_{jk}|
\]
are entrywise norms, and $\norm{A}_F$ is the Frobenius norm.
We use $\svmin(A)$ and $\svmax(A)$ for the smallest and largest
singular values. The following covariance concentration result
is from~\citet{ravikumar2011high}.

\begin{lemma}
\label{lem:covelewise}
For $\bX\sim P^{\otimes n}$ and every $i,j\in[p]$,
\[
 \mathsf{Pr}_{\bX\sim P^{\otimes n}}
 \bigl(\abs{S(\bX)_{ij}-\Sigma_{ij}}>\eps\bigr)
 \le4\exp\left\{
 -\frac{n\eps^2}
 {128(1+4\sigma^2)^2(\max_j\Sigma_{jj})^2}
 \right\}
\]
for $0<\eps<8(1+4\sigma^2)\max_j\Sigma_{jj}$.
\end{lemma}

\begin{proposition}
\label{thm:acv_cv_lossbound}
Let $\bX\sim P^{\otimes n}$ and fix $\lambda>0$.
For every $\delta>0$,
\[
 \mathsf{Pr}_{\bX\sim P^{\otimes n}}
 \bigl(\abs{\ACV(\lambda)-\CV(\lambda)}<\delta\bigr)
 \to1\qquad\text{as }n\to\infty.
\]
\end{proposition}

\begin{proof}[Proof of Proposition~\ref{thm:acv_cv_lossbound}]
We use the following entrywise estimate.

\begin{restatable}{lemma}{estelewise}\label{lem:estelewise}
For every data matrix $\bX\in\mathbb R^{n\times p}$ with $n\ge2$,
every $\lambda>0$, and every $i\in[n]$,
\begin{align*}
 \norm{\wTheta^{\lambda,\bX^{(-i)}}-
              \wTheta^{\lambda,\bX}}_1
 &\le \frac{p^2}{\lambda^2}\norm{S-S_{-i}}_\infty,\\
 \norm{\widetilde{\Theta}^{\lambda,\bX^{(-i)}}-
              \wTheta^{\lambda,\bX}}_1
 &\le \frac{p^2}{\lambda^2}\norm{S-S_{-i}}_\infty,\\
 \norm{\widetilde{\Theta}^{\lambda,\bX^{(-i)}}-
              \wTheta^{\lambda,\bX^{(-i)}}}_\infty
 &\le
 \norm{\widetilde{\Theta}^{\lambda,\bX^{(-i)}}-
              \wTheta^{\lambda,\bX^{(-i)}}}_1\\
 &\le \frac{2p^2}{\lambda^2}\norm{S-S_{-i}}_\infty.
\end{align*}
\end{restatable}
The proof is given in Section~\ref{sec:estelewise}. In particular,
the lemma does not assume that the full-data and leave-one-out
estimators have the same support.

Let $C,c>0$ denote constants
depending only on the standardized marginal sub-Gaussian parameter;
their values may change between displays. Choose $\eps>0$ satisfying
\[
 \eps\le c\min\left\{
       \max_j\Sigma_{jj},
       \frac{\lambda}{p}
       \left(1+\frac{\lambda}{\max_j\Sigma_{jj}}\right)^{-1/2}
       \right\},
 \qquad
 \frac{n\eps^2}{(\max_j\Sigma_{jj})^2}\ge1.
\]
By Lemma~\ref{lem:covelewise}, the marginal sub-Gaussian tail
bound, and a union bound, the simultaneous event
\begin{align*}
 \max_{j,k\in[p]}\abs{S_{jk}-\Sigma_{jk}}&\le\eps,
 &
 \max_{\substack{i\in[n]\\j\in[p]}}
       \bigl(\bX^{(i)}_j\bigr)^2
 &\le\frac{n\eps^2}{\max_j\Sigma_{jj}}
 \numberthis\label{eq:inner_approx1}
\end{align*}
has probability at least
\[
 1-Cnp^2
 \exp\left\{-\frac{cn\eps^2}{(\max_j\Sigma_{jj})^2}\right\}.
\]
Work on this event. Since $S$ and $S_i$ are positive semidefinite,
\[
 \norm{S}_\infty=\max_jS_{jj}\le2\max_j\Sigma_{jj},
 \qquad
 \norm{S_i}_\infty=\max_j\bigl(\bX^{(i)}_j\bigr)^2.
\]
The exact identity $S-S_{-i}=(S_i-S)/(n-1)$ then gives
\[
 \max_{i\in[n]}\norm{S-S_{-i}}_\infty
 \le
 \frac{n\eps^2/(\max_j\Sigma_{jj})+2\max_j\Sigma_{jj}}{n-1}
 \le \frac{6\eps^2}{\max_j\Sigma_{jj}}.
\]

We first verify that all approximate estimators in the score are
positive definite. Because the penalty includes the diagonal
entries, the diagonal KKT identities are
\[
 \bigl[(\wTheta^{\lambda,\bX})^{-1}\bigr]_{jj}=S_{jj}+\lambda,
 \qquad
 \bigl[(\wTheta^{\lambda,\bX^{(-i)}})^{-1}\bigr]_{jj}
       =(S_{-i})_{jj}+\lambda.
\]
For any symmetric matrix $B$, weighted Cauchy--Schwarz gives
\[
 B\succeq
 -\left(\max_j
       \bigl[(\wTheta^{\lambda,\bX})^{-1}\bigr]_{jj}\right)
       \norm{B}_1\,\wTheta^{\lambda,\bX}.
\]
Apply this inequality to
$B=\widetilde{\Theta}^{\lambda,\bX^{(-i)}}-\wTheta^{\lambda,\bX}$
and use Lemma~\ref{lem:estelewise}. It follows that
\begin{align*}
 \widetilde{\Theta}^{\lambda,\bX^{(-i)}}
 &\succeq
 \left[
  1-\frac{p^2(\max_jS_{jj}+\lambda)}{\lambda^2}
       \norm{S-S_{-i}}_\infty
 \right]\wTheta^{\lambda,\bX}\\
 &\succeq \frac12\wTheta^{\lambda,\bX}\succ0.
\end{align*}
For the second inequality, use
\[
 \frac{p^2(\max_jS_{jj}+\lambda)}{\lambda^2}
       \max_i\norm{S-S_{-i}}_\infty
 \le
 \frac{Cp^2}{\lambda^2}
       \left(1+\frac{\lambda}{\max_j\Sigma_{jj}}\right)\eps^2
 \le\frac12,
\]
where the constant $c$ in the restriction on $\eps$ is chosen
sufficiently small.

Positive definiteness and the preceding lower bound imply
\[
 \norm{(\widetilde{\Theta}^{\lambda,\bX^{(-i)}})^{-1}}_\infty
 \le2(\max_jS_{jj}+\lambda).
\]
Indeed, inversion reverses the positive semidefinite order, and
the maximum absolute entry of a positive semidefinite matrix
equals its largest diagonal entry. Similarly,
$(S_{-i})_{jj}\le nS_{jj}/(n-1)\le2\max_jS_{jj}$ gives
\[
 \norm{(\wTheta^{\lambda,\bX^{(-i)}})^{-1}}_\infty
 \le2\max_jS_{jj}+\lambda
 \le2(\max_jS_{jj}+\lambda).
\]

Let
$B_i=\widetilde{\Theta}^{\lambda,\bX^{(-i)}}-
          \wTheta^{\lambda,\bX^{(-i)}}$.
Applying concavity of $\log\det$ in both directions gives
\begin{align*}
 \inner{(\widetilde{\Theta}^{\lambda,\bX^{(-i)}})^{-1}}{B_i}
 &\le
 \log\det\widetilde{\Theta}^{\lambda,\bX^{(-i)}}-
 \log\det\wTheta^{\lambda,\bX^{(-i)}}\\
 &\le
 \inner{(\wTheta^{\lambda,\bX^{(-i)}})^{-1}}{B_i}.
\end{align*}
Consequently,
\begin{align*}
 \abs{\ACV(\lambda)-\CV(\lambda)}
 &\le\frac1n\sum_{i=1}^n
       \left[2(\max_jS_{jj}+\lambda)+\norm{S_i}_\infty\right]
       \norm{B_i}_1\\
 &\le\frac{2p^2}{\lambda^2n}\sum_{i=1}^n
       \left[2(\max_jS_{jj}+\lambda)+\norm{S_i}_\infty\right]
       \norm{S-S_{-i}}_\infty.
 \numberthis\label{eq:acvgap}
\end{align*}
This absolute bound uses both log-determinant inequalities.

The event~\eqref{eq:inner_approx1} also implies
\begin{align*}
 \frac1n\sum_{i=1}^n\norm{S_i}_\infty
 &\le\frac1n\sum_{i=1}^n\Tr(S_i)
   =\Tr(S)\le2p\max_j\Sigma_{jj},\\
 \frac1n\sum_{i=1}^n\norm{S_i}_\infty^2
 &\le\left(\max_i\norm{S_i}_\infty\right)
       \frac1n\sum_{i=1}^n\norm{S_i}_\infty
 \le2np\eps^2.
 \numberthis\label{eq:acvgapterm1}
\end{align*}
Using these estimates and
$\norm{S-S_{-i}}_\infty
 \le(\norm{S_i}_\infty+\max_jS_{jj})/(n-1)$
in~\eqref{eq:acvgap}, we obtain
\begin{align*}
 \abs{\ACV(\lambda)-\CV(\lambda)}
 &\le
 \frac{Cp^3}{\lambda^2}
 \left[
   \eps^2+
   \frac{(\max_j\Sigma_{jj})(\max_j\Sigma_{jj}+\lambda)}{n}
 \right]\\
 &\le
 \frac{Cp^3}{\lambda^2}
 \left(1+\frac{\lambda}{\max_j\Sigma_{jj}}\right)\eps^2,
 \numberthis\label{eq:acvgapterm2}
\end{align*}
where the last inequality uses
$n\eps^2/(\max_j\Sigma_{jj})^2\ge1$.

For each fixed $\lambda>0$, choose
\[
 \eps=
 \frac{c}{p^{3/2}n^{1/20}}
 \min\left\{
    \max_j\Sigma_{jj},
    \frac{\lambda}{\sqrt{1+\lambda/(\max_j\Sigma_{jj})}}
 \right\}.
\]
This satisfies both upper restrictions on $\eps$.
Whenever $n\eps^2/(\max_j\Sigma_{jj})^2\ge1$, the preceding bound gives
\[
 \abs{\ACV(\lambda)-\CV(\lambda)}\le Cn^{-1/10}
\]
with probability at least
\[
 1-Cnp^2\exp\left\{
 -\frac{c\,n^{9/10}}{p^3}
 \min\left\{
  1,\frac{\lambda^2}
  {(\max_j\Sigma_{jj})^2+\lambda\max_j\Sigma_{jj}}
 \right\}
 \right\}.
\]
This displays the polynomial dimension and covariance factors.
In the sufficient-sample joint limit stated at the start of this
appendix, the failure probability tends to zero; in particular,
$n\eps^2/(\max_j\Sigma_{jj})^2\ge1$ eventually holds.
The proposition follows.
\end{proof}

\subsection{Proof of Lemma \ref{lem:acv}}
\label{sec:acv existence}
We shorten $\lambdao^{\bX}$ and $\lambdaacv^{\bX}$ to
$\lambdao$ and $\lambdaacv$ when there is no ambiguity.
For every $\lambda>0$, define
\begin{align*}
 Q(\lambda,\bX)
 &:=
 \E_{Y\sim P}\!\left[
 -\log\det\wTheta^{\lambda,\bX}
 +\Tr(S(Y)\wTheta^{\lambda,\bX})\right]\\
 &=
 -\log\det\wTheta^{\lambda,\bX}
 +\Tr(\Sigma\wTheta^{\lambda,\bX}),\\
 Q_{\bZ}(\lambda,\bX)
 &:=
 -\log\det\wTheta^{\lambda,\bX}
 +\Tr(S(\bZ)\wTheta^{\lambda,\bX}).
\end{align*}
The second definition applies to arbitrary training and evaluation
matrices of compatible dimension.

\begin{proof}[Proof of Lemma~\ref{lem:acv}]
We use the completed positive selection rules stated after the
definition of $\lambdaacv$. Their existence is justified at the
end of this proof. In particular, on samples where the ACV
infimum is finite, for every $\lambda>0$,
\begin{align*}
 \ACV(\lambdaacv)&\le\ACV(\lambda)+n^{-2},\\
 Q(\lambdao,\bX)&\le Q(\lambda,\bX)+n^{-2}.
\end{align*}
These inequalities include exact minimizers.

\paragraph{Concentration and leave-one-out differences.}
Let $C,c>0$ denote constants depending only on the common
standardized marginal sub-Gaussian bound; their values may change
between displays. Take
\[
 0<\eps'\le\frac{c\,\svmin(\Sigma)}p,
 \qquad
 \frac{n(\eps')^2}{(\max_j\Sigma_{jj})^2}\ge1,
\]
with $c$ sufficiently small.
Lemma~\ref{lem:covelewise}, the marginal sub-Gaussian tail bound,
and a union bound show that
\[
 \norm{S-\Sigma}_\infty\le\eps',
 \qquad
 \max_{\substack{i\in[n]\\j\in[p]}}\abs{\bX_{ij}}^2
 \le\frac{n(\eps')^2}{\max_j\Sigma_{jj}}
\]
hold simultaneously with probability at least
\[
 1-Cnp^2\exp\left\{
 -\frac{cn(\eps')^2}{(\max_j\Sigma_{jj})^2}
 \right\}.
\]
Throughout the following argument we work on this event.
Then
\[
 \svmin(S)\ge\tfrac12\svmin(\Sigma),\qquad
 \Tr(S)\le2p\max_j\Sigma_{jj}.
\]
Since $S-S_{-i}=(S_i-S)/(n-1)$ and
$\norm{S_i}_F=\Tr(S_i)$, we have
\begin{align*}
 \max_i\norm{S-S_{-i}}_F
 &\le\frac{Cp(\eps')^2}{\max_j\Sigma_{jj}},\\
 \frac1n\sum_i\norm{S-S_{-i}}_F^2
 &\le\frac{n^{-1}\sum_i\norm{S_i}_F^2}{(n-1)^2}
 \le\frac{Cp^2(\eps')^2}{n}.
\end{align*}
where the last step uses $\frac1n\sum_i\norm{S_i}_F^2
 \le\max_i\Tr(S_i)\Tr(S)
 \le2np^2(\eps')^2$.

\paragraph{A global comparison with empirical risk.}
For each fixed $\lambda$, the approximate update is linear in
$S-S_{-i}$, with the same full-data support for every $i$.
Consequently,
\[
 \frac1n\sum_i\widetilde{\Theta}^{\lambda,\bX^{(-i)}}
 =\wTheta^{\lambda,\bX}.
\]
Furthermore, $\inner{S-S_{-i}}
 {\widetilde{\Theta}^{\lambda,\bX^{(-i)}}-\wTheta^{\lambda,\bX}}
 \ge0$.
Concavity of $\log\det$ and
$S_i=S+(n-1)(S-S_{-i})$ therefore give
\[
 Q_{\bX}(\lambda,\bX)\le\ACV(\lambda).
\]
This also holds for invalid candidates, whose ACV value is
$+\infty$. In particular, $S\succ0$ implies
\[
 \ACV(\lambda)\ge Q_{\bX}(\lambda,\bX)
 \ge p+\log\det S,
\]
Thus the ACV infimum is bounded below on this event; the positive
comparison penalties below show that it is finite.

\paragraph{Positive comparison penalties.}
Consider any
\[
 0<\lambda\le\frac{\svmin(S)}{np}.
\]
This interval is used only for comparison and does not restrict
the tuning search. The KKT conditions imply
\[
 \norm{(\wTheta^{\lambda,\bX})^{-1}-S}_\infty\le\lambda,
 \qquad
 \big[(\wTheta^{\lambda,\bX})^{-1}-S\big]_{jj}=\lambda.
\]
Hence
\[
 \norm{
 S^{-1/2}\big((\wTheta^{\lambda,\bX})^{-1}-S\big)S^{-1/2}
 }_F\le\frac1n.
\]
By the scalar inequality
$(1+t)^{-1}+\log(1+t)-1\le2t^2$ for $|t|\le1/2$, 
\[
 0\le Q_{\bX}(\lambda,\bX)-p-\log\det S\le2n^{-2}.
\]
It also follows that
\[
 \svmax(\wTheta^{\lambda,\bX})\le\frac2{\svmin(S)},
 \qquad
 \svmax((\wTheta^{\lambda,\bX})^{-1})\le2\svmax(S).
\]

The support restriction contracts the Frobenius norm, hence
\begin{align*}
 &\norm{
 (\wTheta^{\lambda,\bX})^{-1/2}
 \big(\widetilde{\Theta}^{\lambda,\bX^{(-i)}}-
                         \wTheta^{\lambda,\bX}\big)
 (\wTheta^{\lambda,\bX})^{-1/2}
 }_F\le
 \frac{8\svmax(S)}{\svmin(S)^2}\norm{S-S_{-i}}_F
 \le C\left(\frac{p\eps'}{\svmin(\Sigma)}\right)^2.
\end{align*}
Here we used $\svmax(\Sigma)\le p\max_j\Sigma_{jj}$.
Choosing $c$ sufficiently small makes the final bound at most
$1/2$, so every approximate fold at these comparison penalties
is positive definite.

To bound the correction, use
$\sum_i(\widetilde{\Theta}^{\lambda,\bX^{(-i)}}-
\wTheta^{\lambda,\bX})=0$ and
$0\le t-\log(1+t)\le t^2$ for $|t|\le1/2$.
The exact difference between the ACV score and empirical risk
therefore satisfies
\begin{align*}
 0\le{}&\ACV(\lambda)-Q_{\bX}(\lambda,\bX)\\
 \le{}&
 \left[
  \frac{4(n-1)}{\svmin(S)^2}
  +\frac{64\svmax(S)^2}{\svmin(S)^4}
 \right]\frac1n\sum_i\norm{S-S_{-i}}_F^2\\
 \le{}&C\left(\frac{p\eps'}{\svmin(\Sigma)}\right)^2.
\end{align*}
The first coefficient bounds the quadratic form of the
support-restricted Kronecker matrix; the second bounds the
quadratic log-determinant remainder. For the last inequality,
the restrictions on $\eps'$ imply
\[
 n\ge\frac{(\max_j\Sigma_{jj})^2}{(\eps')^2}
 \ge\frac{p^2(\max_j\Sigma_{jj})^2}
          {c^2\svmin^2(\Sigma)}
 \ge\frac{\svmax^2(\Sigma)}{c^2\svmin^2(\Sigma)}.
\]
Thus the second coefficient contributes at most a constant
multiple of the first.

The same comparison penalties also satisfy
\[
 \norm{
 \Sigma^{-1/2}
 \big((\wTheta^{\lambda,\bX})^{-1}-\Sigma\big)
 \Sigma^{-1/2}
 }_F
 \le\frac{p\eps'}{\svmin(\Sigma)}+\frac{3}{2n}.
\]
Applying the preceding scalar likelihood inequality now gives
\[
 0\le Q(\lambda,\bX)-p-\log\det\Sigma
 \le C\left[
       \left(\frac{p\eps'}{\svmin(\Sigma)}\right)^2+n^{-2}
      \right].
\]

\paragraph{Risk comparison at the selected penalties.}
Choose a sufficiently large constant in
\[
 \eps_n:=
 C\left[
       \left(\frac{p\eps'}{\svmin(\Sigma)}\right)^2+n^{-2}
   \right].
\]
By the selection inequalities and the preceding comparisons,
\begin{align*}
 0\le Q_{\bX}(\lambdaacv,\bX)-p-\log\det S&\le\eps_n,\\
 0\le Q(\lambdao,\bX)-p-\log\det\Sigma&\le\eps_n.
\end{align*}
Taking $c$ sufficiently small makes these gaps at most $1/8$, and the scalar function $t-\log t-1$ exceeds $1/8$ outside
$[1/2,2]$. Applied to the corresponding relative covariance
matrices, this implies
\[
 \svmax(\wTheta^{\lambdaacv,\bX})\le\frac4{\svmin(\Sigma)},
 \qquad
 \svmax(\wTheta^{\lambdao,\bX})\le\frac2{\svmin(\Sigma)}.
\]

The centered comparison follows
\begin{align*}
 -n^{-2}\le{}&
 Q(\lambdaacv,\bX)-Q(\lambdao,\bX)\\
 \le{}&\eps_n+
 \Tr\!\left[
   (\Sigma-S)
   (\wTheta^{\lambdaacv,\bX}-\wTheta^{\lambdao,\bX})
 \right]\\
 \le{}&\eps_n+
 p\eps'\norm{\wTheta^{\lambdaacv,\bX}-\wTheta^{\lambdao,\bX}}_F .
\end{align*}
On the segments from $\Sigma^{-1}$ to the two selected estimates,
the Hessian of $-\log\det\Theta+\Tr(\Sigma\Theta)$ has
Frobenius curvature at least $\svmin^2(\Sigma)/16$, and its gradient vanishes at $\Sigma^{-1}$. By Taylor's theorem,
\begin{align*}
 &\norm{\wTheta^{\lambdaacv,\bX}-\wTheta^{\lambdao,\bX}}_F\\
 &\quad\le\frac{4\sqrt2}{\svmin(\Sigma)}
 \left[
   \sqrt{Q(\lambdaacv,\bX)-p-\log\det\Sigma}
   +\sqrt{Q(\lambdao,\bX)-p-\log\det\Sigma}
 \right].
\end{align*}
Substituting into the centered comparison and applying Young's
inequality yields
\begin{align*}
 &Q(\lambdaacv,\bX)-p-\log\det\Sigma\\
 &\quad\le
 3\big[Q(\lambdao,\bX)-p-\log\det\Sigma\big]
 +2\eps_n+\frac{64p^2(\eps')^2}{\svmin^2(\Sigma)}\\
 &\quad\le C\left[
        \left(\frac{p\eps'}{\svmin(\Sigma)}\right)^2+n^{-2}
       \right].
\end{align*}
This establishes the desired population-risk comparison.

\paragraph{From risk to the penalty difference.}
For any $\Theta\succ0$ with
$-\log\det\Theta+\Tr(\Sigma\Theta)-p-\log\det\Sigma\le1/8$,
\[
 \abs{(\Theta^{-1}-\Sigma)_{jj}}
 \le\Sigma_{jj}
 \sqrt{32\big[-\log\det\Theta+\Tr(\Sigma\Theta)
                         -p-\log\det\Sigma\big]}.
\]
The diagonal KKT identities for the two full-data fits imply
\[
 \lambdaacv-\lambdao
 =
 \big[(\wTheta^{\lambdaacv,\bX})^{-1}
     -(\wTheta^{\lambdao,\bX})^{-1}\big]_{jj}.
\]
Applying the preceding inequality to both fits gives
\[
 \abs{\lambdaacv-\lambdao}
 \le C\max_j\Sigma_{jj}
       \left(\frac{p\eps'}{\svmin(\Sigma)}+\frac1n\right).
\]

Choose $\eps'=
 \frac{c\,\svmin(\Sigma)}
 {p(1+\max_j\Sigma_{jj})n^{1/10}}$. This automatically satisfies the upper restriction on $\eps'$. Whenever $n(\eps')^2/(\max_j\Sigma_{jj})^2\ge1$, it also implies
$\max_j\Sigma_{jj}/n\le Cn^{-1/5}$, and therefore
\[
 \abs{\lambdaacv-\lambdao}\le Cn^{-1/10}
\]
with probability at least
\[
 1-Cnp^2\exp\left\{
 -\frac{c\,n^{4/5}\svmin^2(\Sigma)}
 {p^2(1+\max_j\Sigma_{jj})^2(\max_j\Sigma_{jj})^2}
 \right\}.
\]
Similar to previous arguments, it holds with probability tending to 1.

\paragraph{Existence of the tuning selections.}
For every $\lambda>0$, strict convexity, the log-determinant barrier, and the full-entry penalty ensure a unique fit. The oracle objective is finite and bounded below by $p+\log\det\Sigma$, so its infimum is finite.

At $\lambda=1+\max_jS_{jj}$, the KKT conditions give diagonal approximate folds with
\[
 \widetilde{\Theta}^{\lambda,\bX^{(-i)}}_{jj}
 =\frac{\lambda+\frac{n-2}{n-1}S_{jj}
                   +\frac1{n-1}(S_i)_{jj}}
 {(S_{jj}+\lambda)^2}>0.
\]
Thus ACV has a finite feasible value, and when $S\succ0$ its infimum is bounded below by $p+\log\det S$. Finite infima admit positive $n^{-2}$-optimal choices, and the prescribed fallback handles an ACV infimum of $-\infty$.

The KKT solution map and fold-validity sets are semialgebraic. Both scores are therefore jointly Borel and piecewise continuous in $\lambda$, with finitely many pieces and $\sigma$-compact sublevel and selection sections. The Borel projection and selection theorem yields measurable infima and exact or $n^{-2}$-optimal selections \citep[][]{moschovakis2009descriptive}.
The fallback is measurable as well.

\end{proof}

\section{Evaluation on graph recovery}
In this section, we assess the performance of $\lambdaacv$ in structure learning by evaluating its ability to learn true edges. We divide the analysis into two parts: we first show $\lambdaacv$ is able to locate all true edges asymptotically by proving Theorem \ref{thm:acv overselect} in Section \ref{sec:acv overselect}. We next show that $\lambdaacv$ alone fails the consistency by proving Theorem \ref{thm:inconsis_G} in Section \ref{sec:inconsis_G}. We drop the dependence on $\bX$ if there is no ambiguity.

We first recall the KKT conditions~\citep[e.g.][]{ravikumar2011high}:
\begin{restatable}{lemma}{kkt}\label{lem:kkt}
The convex program in~\eqref{eq:obj func} has a unique optimizer satisfying the KKT conditions 
    \begin{align*}
    \numberthis\label{eq:block_kkt}
        \begin{cases}
            (\wTheta^{\lambda})^{-1}_{ij} - S_{ij} = \lambda\cdot\sign(\wTheta^{\lambda}_{ij}) & \wTheta^{\lambda}_{ij}\neq 0\\
            \abs{(\wTheta^{\lambda})^{-1}_{ij} - S_{ij}}\leq\lambda & \wTheta^{\lambda}_{ij}= 0
        \end{cases}
    \end{align*}
    for any $\lambda>0$. 
\end{restatable}

\subsection{Proof of Theorem \ref{thm:acv overselect}}\label{sec:acv overselect}
\begin{proof}
If $E=\varnothing$, the conclusion is immediate. Otherwise, for
every $i\neq j$,
\[
 \inf_{\substack{\Theta\succ0\\\Theta_{ij}=0}}
 \{-\log\det\Theta+\Tr(\Sigma\Theta)\}
 =p+\log\det\Sigma-\log(1-\rho_{ij\mid G}^{\,2}).
\]
To verify this identity, modify only the $(i,j)$ and $(j,i)$
entries of $\Sigma$ so that the off-diagonal entry of the
corresponding $2\times2$ Schur complement is zero.
Its diagonal entries remain unchanged and positive, so the
modified matrix is positive definite. Its inverse has zero
$(i,j)$ entry and satisfies the constrained first-order equations,
and hence is the unique minimizer. The modified matrix has
determinant $\det\Sigma/(1-\rho_{ij\mid G}^{\,2})$, and the
trace term at this minimizer equals $p$, proving the identity.

Consequently, if the ACV-selected fit omits a true edge $(i,j)$,
then (A3) gives
\begin{align*}
 Q(\lambdaacv,\bX)-p-\log\det\Sigma
 &\ge -\log(1-\rho_{ij\mid G}^{\,2})\\
 &\ge \rho_{ij\mid G}^{\,2}
 \ge C_3^2n^{-2\gamma}.
\end{align*}

The proof of Lemma~\ref{lem:acv} in Appendix~A.3 establishes
\[
 Q(\lambdaacv,\bX)-p-\log\det\Sigma
 \le C\left[
 \left(\frac{p\eps'}{\svmin(\Sigma)}\right)^2+n^{-2}
 \right]
\]
with probability at least
\[
 1-Cnp^2\exp\left\{
 -\frac{cn(\eps')^2}{(\max_j\Sigma_{jj})^2}
 \right\},
\]
provided $0<\eps'\le c\,\svmin(\Sigma)/p$ and
$n(\eps')^2/(\max_j\Sigma_{jj})^2\ge1$.
Choose
\[
 \eps'=\frac{c C_3\svmin(\Sigma)}{p n^\gamma}
\]
with $c>0$ sufficiently small.
Since $C_3n^{-\gamma}\le\min_{(i,j)\in E}|\rho_{ij\mid G}|<1$,
the upper restriction on $\eps'$ holds. The other restriction
holds eventually in the sufficient-sample regime below.
As $n^{-2}=o(n^{-2\gamma})$, the risk bound is at most
$\tfrac12 C_3^2n^{-2\gamma}$ for sufficiently large $n$.
This excludes every missing true edge on the same event. Thus
\[
 \mathbf{Pr}_{\bX\sim P^{\otimes n}}
 \left(G\not\subseteq\widehat G^{\lambdaacv,\bX}\right)
 \le Cnp^2\exp\left\{
 -\frac{c C_3^2 n^{1-2\gamma}\svmin^2(\Sigma)}
 {p^2(\max_j\Sigma_{jj})^2}
 \right\}
 \longrightarrow0
\]
under the global sufficient-sample convention.
In particular, under (A2), a sufficient condition for this last
bound to vanish is $p^2\log(np)=o(n^{1-2\gamma})$.
\end{proof}

\subsection{Proof of Theorem \ref{thm:inconsis_G}}\label{sec:inconsis_G}
Use $S=S(\bX)$ and $S_i=S(\bX^{(i)})$ as in Appendix~\ref{sec:app A}.
For $n\ge p$ and $n>2$, Gaussian sampling implies that $S\succ0$ and every entry of $S^{-1}$ is nonzero almost surely. Indeed, $S$ has a density on the positive-definite cone, and each off-diagonal equation $(S^{-1})_{ij}=0$ defines a measure-zero algebraic set. For each such sample, the KKT conditions give
\[
 \|(\widehat\Theta^{\lambda,\bX})^{-1}-S\|_\infty
 \le\lambda,
\]
so
$\widehat\Theta^{\lambda,\bX}\to S^{-1}$ as $\lambda\downarrow0$.
Consequently, its support is full for all sufficiently small
positive $\lambda$. By~\eqref{eq:acv},
\[
 \lim_{\lambda\downarrow0}
 \widetilde\Theta^{\lambda,\bX^{(-i)}}
 =
 \frac{n-2}{n-1}S^{-1}
 +\frac1{n-1}S^{-1}S_iS^{-1}
 \succ0.
\]
Thus all approximate folds are valid for sufficiently small positive $\lambda$, and the limit of $\ACV(\lambda)$ is finite. 

The matrix determinant lemma gives
\begin{align*}
 \MoveEqLeft\lim_{\lambda\downarrow0}\ACV(\lambda)-p-\log\det S\\
 &=
 -p\log\!\left(\frac{n-2}{n-1}\right)
 -\frac1n\sum_{i=1}^n
   \log\!\left(1+\frac{\Tr(S_iS^{-1})}{n-2}\right)\\
 &\quad-\frac{p}{n-1}
 +\frac1{n(n-1)}
   \sum_{i=1}^n\{\Tr(S_iS^{-1})\}^2.
\end{align*}
Here
\[
 \frac1n\sum_{i=1}^n\Tr(S_iS^{-1})=p,
 \qquad
 \frac1n\sum_{i=1}^n\{\Tr(S_iS^{-1})\}^2
 \xrightarrow{}p(p+2).
\]
The second assertion follows from $S^{-1}\to\Sigma^{-1}$ and
the law of large numbers for Gaussian fourth moments, since
$\E[(X^\top\Sigma^{-1}X)^2]=p(p+2)$.
Using $0\le u-\log(1+u)\le u^2/2$ for $u\ge0$ therefore yields
\[
 n\left\{
 \lim_{\lambda\downarrow0}\ACV(\lambda)-p-\log\det S
 \right\}
 \xrightarrow{}p(p+1).
\]
This limit is used only to compare scores over positive
penalties; it does not assert any limit for $\lambdaacv^{\bX}$.

Appendix~\ref{sec:acv existence} gives
$Q_{\bX}(\lambda,\bX)\le\ACV(\lambda)$ for every $\lambda>0$.
Since the ACV infimum is finite when $S\succ0$, the stated
selection convention also gives
\[
 \ACV(\lambdaacv^{\bX})
 \le\inf_{\lambda>0}\ACV(\lambda)+n^{-2}
 \le\lim_{\lambda\downarrow0}\ACV(\lambda)+n^{-2}.
\]
On the event $\{\widehat G^{\lambdaacv^{\bX},\bX}=G\}$,
the selected fit satisfies all zero constraints of $G$.
Consequently,
\begin{align*}
 \inf_{\substack{\Theta\succ0\\
          \Theta_{ij}=0\ \text{if }i<j,\ (i,j)\notin E}}
   \{-\log\det\Theta+\Tr(S\Theta)\}
 &\le
 Q_{\bX}(\lambdaacv^{\bX},\bX)
 \le\ACV(\lambdaacv^{\bX})\\
 &\le\lim_{\lambda\downarrow0}\ACV(\lambda)+n^{-2}.
\end{align*}

The Gaussian model with these zero constraints is a regular
model of dimension $p+|E|$, nested in the unrestricted model
of dimension $p(p+1)/2$. The true precision matrix lies in its
relative interior, and the Fisher information is positive definite
on the free coordinates. Wilks' theorem~\citep[][]{wilks1938large} gives
\[
 n\left[
 \inf_{\substack{\Theta\succ0\\
          \Theta_{ij}=0\ \text{if }i<j,\ (i,j)\notin E}}
   \{-\log\det\Theta+\Tr(S\Theta)\}
 -p-\log\det S
 \right]
 \xrightarrow{d}\chi^2_{\binom{p}{2}-|E|}.
\]
The factor $n$ is the likelihood-ratio normalization because
the Gaussian log likelihood is $-n/2$ times the displayed
empirical objective, up to an additive constant.
Combining the last three components and using continuity of
the $\chi^2$-distribution gives
\[
 \limsup_{n\to\infty}
 \mathbf{Pr}_{\bX\sim N(0,\Sigma)^{\otimes n}}
 \left(\widehat G^{\lambdaacv^{\bX},\bX}=G\right)
 \le
 \mathbf{Pr}\!\left(
   \chi^2_{\binom{p}{2}-|E|}\le p(p+1)
 \right)<1,
\]
where the strict inequality follows from
$\binom{p}{2}-|E|\ge1$.

\section{FDR control and consistency}
In this section, we prove false-positive control and selection consistency for
Algorithm~\ref{alg: post cleaning}. The global asymptotic convention
and the final failure bound in Appendix~\ref{sec:acv existence} imply
\[
 p^2\log(np)=O(n^{4/5}),\qquad
 \frac{p\{\log(np)\}^2}{\sqrt n}\longrightarrow0,\qquad
 \frac{p^2}{n}\longrightarrow0.
\]
We use these consequences throughout this appendix and the proofs of its supporting results. We shorten our notation $ \wTheta^{\lambda^\bX,\bX}, S(\bX), \lambdaacv^{\bX}$ to $\wTheta^{\lambda}, S, \lambdaacv$ if there is no ambiguity on $\bX\sim P^{\otimes n}$.

\subsection{Proof of Theorem \ref{thm:fdr control}}\label{sec:fdr control pf}
Let
\[
 \mathcal H_0:=
 \{(i,j):1\le i<j\le p,\ (i,j)\notin E\}.
\]
For analysis, extend the formulas for $U(i,j)$ and $\widehat T_{ij}$ in Algorithm~\ref{alg: post cleaning} to all pairs, including those not tested by the algorithm. We use the following uniform null approximation, proved in
Section~\ref{sec:Tstat}. 

\begin{restatable}{proposition}{Tstat}
\label{prop:Tstat}
Let $\bX\sim P^{\otimes n}$ and assume Assumptions \ref{asump:high dim},~\ref{asump:nonsingular} and~\ref{asump:indep}. Let $\widehat G$ be a measurable function of $\bX$ such that
$G\subseteq\widehat G$ almost surely.
Then, for every fixed $M>0$,
\[
 \sup_{(i,j)\in\mathcal H_0}
 \sup_{0\le t\le M\sqrt{\log(np)}}
 \left|
 \frac{
  \mathbf{Pr}_{\bX\sim P^{\otimes n}}
       (|\widehat T_{ij}|\ge t)
 }{2\{1-\Phi(t)\}}-1
 \right|
 \longrightarrow0.
\]
If $\mathcal H_0$ is empty, the assertion is vacuous.
\end{restatable}

Set $t_\alpha=\Phi^{-1}(1-\alpha/(2p^2))$.
For fixed $\alpha\in(0,1)$,
$t_\alpha=O(\sqrt{\log(np)})$.
Under the containment premise of Theorem~5.2, a union bound over
the deterministic set $\mathcal H_0$ and
Proposition~\ref{prop:Tstat} give
\begin{align*}
 &\mathbf{Pr}_{\bX\sim P^{\otimes n}}
      (\check G^{\alpha,\bX}\not\subseteq G)\\
 &\quad\le
 \sum_{(i,j)\in\mathcal H_0}
 \mathbf{Pr}_{\bX\sim P^{\otimes n}}
      (|\widehat T_{ij}|\ge t_\alpha)\\
 &\quad\le
 |\mathcal H_0|\,\frac{\alpha}{p^2}\{1+o(1)\}
 \le\alpha\{1+o(1)\}.
\end{align*}
Taking the limit superior proves Theorem~5.2.

For an input graph that contains $G$ only with probability tending
to one, apply the same argument to the hypothetical input
$\widehat G\cup G$.
Its output agrees with the actual output on
$\{G\subseteq\widehat G\}$, and therefore
\begin{align*}
 \mathbf{Pr}_{\bX\sim P^{\otimes n}}
      (\check G^{\alpha,\bX}\not\subseteq G)
 &\le
 \mathbf{Pr}_{\bX\sim P^{\otimes n}}
      (G\not\subseteq\widehat G)
 +\alpha\{1+o(1)\}.
\end{align*}
The same argument applies to $\alpha=n^{-\varepsilon}$ for every
fixed $\varepsilon>0$, since then
$t_\alpha=O(\sqrt{\log(np)})$.

\subsection{Proof of Theorem \ref{thm:two_stage}}\label{sec:two_stage}

We first establish simultaneous bounds for the fitted residuals used by Algorithm~\ref{alg: post cleaning}.

\begin{restatable}{lemma}{rhoij}
    \label{lem:rhoij}
        Let $\bX\sim P^{\otimes n}$ and assume \ref{asump:high dim},~\ref{asump:nonsingular} and~\ref{asump:indep}. For each $i\neq j$, let $e_i^0,e_j^0$ be the population residuals from linear regression of $X_i,X_j$ on $X_{[p]\setminus\{i,j\}}$. For every conditioning set
\[
 \bigl(\operatorname{ne}(i)\cup\operatorname{ne}(j)\bigr)
 \setminus\{i,j\}
 \subseteq U(i,j)\subseteq[p]\setminus\{i,j\},
\]
let $e_i,e_j$ be the fitted ordinary least-squares residuals on
$X_{U(i,j)}$, using the same intercept convention as the algorithm,
and write
\[
 \widehat r_{ij}=\frac1n\sum_{k=1}^n e_i^{(k)}e_j^{(k)},
 \qquad
 \wtheta_{ij}=\frac1n\sum_{k=1}^n
 (e_i^{(k)}e_j^{(k)}-\widehat r_{ij})^2.
\]
For sufficiently large $n$, with probability at least $1-Cp^2/n$,
simultaneously over all such pairs and conditioning sets,
\begin{align*}
 \left|\widehat r_{ij}-\E[e_i^0e_j^0]\right|
 &\le C\svmin(\Sigma)\left\{
 \sqrt{\frac{\log(np)}n}+\frac{p\log(np)}n\right\},\\
 c\svmin(\Sigma)
 &\le\sqrt{\wtheta_{ij}}\le C\svmin(\Sigma).
\end{align*}
Here $c,C>0$ depend only on the common distributional bounds.
Moreover,
\[
 \E[e_i^0e_j^0]
 =\rho_{ij\mid G}
 \sqrt{\Var(e_i^0)\Var(e_j^0)},
 \qquad
 \svmin(\Sigma)\le\Var(e_i^0),\Var(e_j^0)
 \le\svmax(\Sigma).
\]
    \end{restatable}
     The proof of this lemma can be found in Section \ref{sec:rhoij}.

Set $\alpha=n^{-\varepsilon}$ for a fixed $\varepsilon>2\tau$
and take $\widehat G=\widehat G^{\lambdaacv,\bX}$ as input to
Algorithm~\ref{alg: post cleaning}.
Write its output as $\check G^{\alpha,\bX}$, as in Appendix~\ref{sec:fdr control pf}.
The bound proved in Appendix~\ref{sec:fdr control pf} and Theorem~\ref{thm:acv overselect} give
\[
 \mathbf{Pr}_{\bX\sim P^{\otimes n}}
 (\check G^{\alpha,\bX}\not\subseteq G)
 \le
 \mathbf{Pr}_{\bX\sim P^{\otimes n}}
 (G\not\subseteq\widehat G)
 +\alpha\{1+o(1)\}
 \longrightarrow0.
\]
If $E=\varnothing$, this proves the theorem. Otherwise, it remains
to show that every true edge is retained.

On $\{G\subseteq\widehat G\}$, every true edge is tested and its
conditioning set satisfies the preceding lemma.
By \ref{asump:signal strength} and the lemma's population identities,
\[
 \min_{(i,j)\in E}|\E[e_i^0e_j^0]|
 \ge C_3\svmin(\Sigma)n^{-\gamma}.
\]
The global asymptotic convention and Appendix~\ref{sec:acv existence} give
$p^2\log(np)=O(n^{4/5})$. Together with \ref{asump:high dim} and
$0\le\gamma<1/2$, this implies
\begin{align*}
 n^\gamma\left\{
 \sqrt{\frac{\log(np)}n}+\frac{p\log(np)}n\right\}
 &\le C\left(n^{\gamma-1/2}+n^{\gamma-3/5}\right)
 \sqrt{\log(np)}\longrightarrow0.
\end{align*}
Consequently, on the simultaneous event of the lemma, for all
sufficiently large $n$,
\[
 \min_{(i,j)\in E}|\widehat r_{ij}|
 \ge\frac{C_3}{2}\svmin(\Sigma)n^{-\gamma},
 \qquad
 \min_{(i,j)\in E}|\widehat T_{ij}|
 \ge c n^{1/2-\gamma}.
\]
The second inequality uses the positive lower bound and the
upper bound for $\sqrt{\wtheta_{ij}}$ in the lemma. The testing threshold satisfies
\[
 t_\alpha=\Phi^{-1}\!\left(1-\frac{\alpha}{2p^2}\right)
 \le\sqrt{2\log(2p^2/\alpha)}
 =O(\sqrt{\log n})
 =o(n^{1/2-\gamma}).
\]
Therefore every true edge is retained on the intersection of the
screening event and the simultaneous event of the lemma. Hence
\[
 \mathbf{Pr}_{\bX\sim P^{\otimes n}}
 (G\not\subseteq\check G^{\alpha,\bX})
 \le
 \mathbf{Pr}_{\bX\sim P^{\otimes n}}
 (G\not\subseteq\widehat G)
 +\frac{Cp^2}{n}
 \longrightarrow0,
\]
where $p^2/n\to0$ follows from the same Appendix~\ref{sec:acv existence} bound. Combining the two error probabilities proves the theorem.

\section{Technical lemmas}
\label{sec:omitted_proof}
\subsection{Proof of Lemma \ref{lem:estelewise}}\label{sec:estelewise}
\estelewise*
\begin{proof}
    Recall that $\norm{\cdot}_1$ denotes the entrywise $\ell_1$ norm.
    By the KKT conditions,
    \begin{align*}
        \Tr\left(S\wTheta^{\lambda,\bX}\right)
        + \lambda\norm{\wTheta^{\lambda,\bX}}_1
        &= p,\\
        \Tr\left(S_{-i}\wTheta^{\lambda,\bX^{(-i)}}\right)
        + \lambda\norm{\wTheta^{\lambda,\bX^{(-i)}}}_1
        &= p.
    \end{align*}
    Since $S,S_{-i}\succeq 0$ and both estimators are positive definite,
    \begin{align*}
        \norm{\wTheta^{\lambda,\bX}}_1
        \leq \frac{p}{\lambda},
        \qquad
        \norm{\wTheta^{\lambda,\bX^{(-i)}}}_1
        \leq \frac{p}{\lambda}.
    \end{align*}

    By weighted Cauchy--Schwarz,
    \begin{align*}
        \MoveEqLeft\norm{
            \wTheta^{\lambda,\bX^{(-i)}}
            - \wTheta^{\lambda,\bX}
        }_1^2\\
        &\quad\leq
        \norm{\wTheta^{\lambda,\bX}}_1
        \norm{\wTheta^{\lambda,\bX^{(-i)}}}_1
        \times
        \Tr\left[
            \left(
                (\wTheta^{\lambda,\bX})^{-1}
                - (\wTheta^{\lambda,\bX^{(-i)}})^{-1}
            \right)
            \left(
                \wTheta^{\lambda,\bX^{(-i)}}
                - \wTheta^{\lambda,\bX}
            \right)
        \right].
    \end{align*}

    Moreover, the KKT conditions and monotonicity of the
    $\ell_1$ subgradient imply
    \begin{align*}
        &\Tr\left[
            \left(
                (\wTheta^{\lambda,\bX})^{-1}
                - (\wTheta^{\lambda,\bX^{(-i)}})^{-1}
            \right)
            \left(
                \wTheta^{\lambda,\bX^{(-i)}}
                - \wTheta^{\lambda,\bX}
            \right)
        \right]\\
        &\quad\leq
        \Tr\left[
            (S-S_{-i})
            \left(
                \wTheta^{\lambda,\bX^{(-i)}}
                - \wTheta^{\lambda,\bX}
            \right)
        \right]\\
        &\quad\leq
        \norm{S-S_{-i}}_\infty
        \norm{
            \wTheta^{\lambda,\bX^{(-i)}}
            - \wTheta^{\lambda,\bX}
        }_1.
    \end{align*}
    Combining the preceding inequalities, we obtain
    \begin{align*}
        \norm{
            \wTheta^{\lambda,\bX^{(-i)}}
            - \wTheta^{\lambda,\bX}
        }_1^2
        &\leq
        \frac{p^2}{\lambda^2}
        \norm{S-S_{-i}}_\infty
        \norm{
            \wTheta^{\lambda,\bX^{(-i)}}
            - \wTheta^{\lambda,\bX}
        }_1.
    \end{align*}
    Cancelling the last norm when it is nonzero, with the zero
    case being immediate, gives
    \begin{align*}
        \norm{
            \wTheta^{\lambda,\bX^{(-i)}}
            - \wTheta^{\lambda,\bX}
        }_\infty
        &\leq
        \norm{
            \wTheta^{\lambda,\bX^{(-i)}}
            - \wTheta^{\lambda,\bX}
        }_1\\
        &\leq
        \frac{p^2}{\lambda^2}\norm{S-S_{-i}}_\infty.
        \numberthis\label{eq:estelewiseproof1}
    \end{align*}
For the approximate update, sum the absolute values over all
coordinates in $\widehat{\mathcal S}$. By its definition,
\begin{align*}
 &\norm{\widetilde{\Theta}^{\lambda,\bX^{(-i)}}-
                 \wTheta^{\lambda,\bX}}_1\\
 &\quad=
 \norm{
  \left(\wTheta^{\lambda,\bX}\otimes\wTheta^{\lambda,\bX}\right)
       _{\widehat{\mathcal S}\widehat{\mathcal S}}
  (S-S_{-i})_{\widehat{\mathcal S}}
 }_1\\
 &\quad\le
 \left[
  \sum_{j\in\widehat{\mathcal S}}
  \sum_{k\in\widehat{\mathcal S}}
  \abs{
   \left(
    \left(\wTheta^{\lambda,\bX}\otimes\wTheta^{\lambda,\bX}\right)
       _{\widehat{\mathcal S}\widehat{\mathcal S}}
   \right)_{jk}
  }
 \right]\norm{S-S_{-i}}_\infty\\
 &\quad\le
 \norm{\wTheta^{\lambda,\bX}\otimes\wTheta^{\lambda,\bX}}_1
       \norm{S-S_{-i}}_\infty\\
 &\quad=
 \norm{\wTheta^{\lambda,\bX}}_1^2\norm{S-S_{-i}}_\infty
 \le\frac{p^2}{\lambda^2}\norm{S-S_{-i}}_\infty.
\end{align*}
Together with the preceding bound for
$\norm{\wTheta^{\lambda,\bX^{(-i)}}-\wTheta^{\lambda,\bX}}_1$,
the triangle inequality gives
\begin{align*}
 \norm{\widetilde{\Theta}^{\lambda,\bX^{(-i)}}-
               \wTheta^{\lambda,\bX^{(-i)}}}_\infty
 &\le
 \norm{\widetilde{\Theta}^{\lambda,\bX^{(-i)}}-
               \wTheta^{\lambda,\bX^{(-i)}}}_1\\
 &\le
 \norm{\widetilde{\Theta}^{\lambda,\bX^{(-i)}}-
               \wTheta^{\lambda,\bX}}_1
 +\norm{\wTheta^{\lambda,\bX}-
               \wTheta^{\lambda,\bX^{(-i)}}}_1\\
 &\le\frac{2p^2}{\lambda^2}\norm{S-S_{-i}}_\infty.
\end{align*}
This proves the lemma.

\end{proof}

\subsection{Proof of Proposition \ref{prop:Tstat}}
\label{sec:Tstat}
\Tstat*
\begin{proof}
Fix $M>0$ and suppose $\mathcal H_0$ is nonempty.
The global asymptotic convention and the final bound in Appendix~\ref{sec:acv existence} imply $p^2\log(np)=O(n^{4/5})$, and hence
\[
 \frac{p\{\log(np)\}^2}{\sqrt n}\longrightarrow0.
\]
For each fixed conditioning set, common rescaling leaves the test
statistic unchanged. Since the bounds below hold uniformly over
these sets,  WLOG we work with $\svmin(\Sigma)=1$ and
$\svmax(\Sigma)\le C_2$.

Let $\varepsilon_i$ be the population error in
Assumption~\ref{asump:indep}. The local Markov property together give $\varepsilon_i\perp X_{[p]\setminus\{i\}}$.
For a null pair, $\varepsilon_j$ is a function of
$X_{[p]\setminus\{i\}}$, so $\varepsilon_i\perp\varepsilon_j$.
Consequently,
\[
 \theta_{ij}:=\Var(\varepsilon_i\varepsilon_j)
 =\Var(\varepsilon_i)\Var(\varepsilon_j),
 \qquad 1\le\theta_{ij}\le C_2^2.
\]
Here $1\le\Var(\varepsilon_i)\le\Sigma_{ii}\le C_2$ follows
from the covariance bounds and regression orthogonality.
Independence and centering also imply
$\E e^{t\varepsilon_i}\le\E e^{tX_i}$, so the population errors
have uniformly bounded sub-Gaussian parameters.

We first control the empirical quantities formed from these errors.
All sample sums below run from $1$ to $n$, and generic constants
$C_M$ depend only on $M$ and the common distributional bounds.
Let $A_1$ be the event on which
\begin{align*}
 \max_{i,j}|S_{ij}-\Sigma_{ij}|
 &\le C_M\sqrt{\frac{\log(np)}n},\\
 \max_i\left(
 \left|\frac1n\sum_k X_i^{(k)}\right|
 +\left|\frac1n\sum_k\varepsilon_i^{(k)}\right|\right)
 &\le C_M\sqrt{\frac{\log(np)}n},\\
 \max_{i,\,\ell\neq i}
 \left|\frac1n\sum_kX_\ell^{(k)}\varepsilon_i^{(k)}\right|
 &\le C_M\sqrt{\frac{\log(np)}n},\\
 \max_{i,k}(|X_i^{(k)}|+|\varepsilon_i^{(k)}|)
 &\le C_M\sqrt{\log(np)},
\end{align*}
and, for every $(i,j)\in\mathcal H_0$,
\[
 \frac{\left|n^{-1}\sum_k
       \varepsilon_i^{(k)}\varepsilon_j^{(k)}\right|}
      {\sqrt{\theta_{ij}}}
 +\left|\frac{\sum_k
       (\varepsilon_i^{(k)}\varepsilon_j^{(k)})^2}
       {n\theta_{ij}}-1\right|
 \le C_M\sqrt{\frac{\log(np)}n}.
\]
Write $A_1^c$ for its complement.
Sub-Gaussian tails and Bernstein's inequality control the covariance
entries, sample means, scores, maxima, and population-product means.
For the squared population products, the tail bound
$\mathbf{Pr}((\varepsilon_i\varepsilon_j)^2>x)
\le C\exp(-c\sqrt{x})$
allows truncation at $C_M\{\log(np)\}^2$.
Bernstein's inequality then gives a deviation of order
$\sqrt{\log(np)/n}+\{\log(np)\}^3/n$.
The second term of the inequality is absorbed by the first under \ref{asump:high dim}, while the
truncation probability and bias can be made arbitrarily small. Choosing the constants so each failure
probability is at most $(np)^{-(M+1)^2-6}$ and taking a union
bound over at most $Cnp^2$ terms yields
\[
 \mathbf{Pr}(A_1^c)\le C_M(np)^{-(M+1)^2-4}.
\]

On $A_1$, the full empirical Gram matrix has smallest eigenvalue
at least $1-C_Mp\sqrt{\log(np)/n}\ge1/2$ for large $n$.
The sample-mean bounds give the same conclusion with an intercept
column adjoined. Every regression Gram matrix is a principal
submatrix of this full matrix, so the inverse bound holds
simultaneously for every conditioning set.
For any null pair and any
$\operatorname{ne}(i)\cup\operatorname{ne}(j)
\subseteq U(i,j)\subseteq[p]\setminus\{i,j\}$,
the population predictors lie in the regression space.
Thus $\varepsilon_i-e_i$ is the ordinary least-square projection of the sampled
population-error vector $\varepsilon_i$. The normal equations give
\begin{align*}
 \frac{\|\varepsilon_i-e_i\|_2^2}{n}
 &\le2\left\{
 \left|\frac1n\sum_k\varepsilon_i^{(k)}\right|^2
 +\sum_{\ell\in U(i,j)}
 \left|\frac1n\sum_kX_\ell^{(k)}\varepsilon_i^{(k)}\right|^2
 \right\}
 \le C_M\frac{p\log(np)}n,\\
 \max_k|\varepsilon_i^{(k)}-e_i^{(k)}|
 &\le C_M\sqrt{p\log(np)}
       \sqrt{\frac{p\log(np)}n}
 =C_M\frac{p\log(np)}{\sqrt n}.
\end{align*}
The sample-mean term covers an intercept and may be omitted
otherwise; the second line also uses the bound on each design row. We note the same estimates hold for arbitrary $j\in[p]$.

Write $\widehat r_{ij}=n^{-1}\sum_k e_i^{(k)}e_j^{(k)}$ and
$\wtheta_{ij}=n^{-1}\sum_k
(e_i^{(k)}e_j^{(k)}-\widehat r_{ij})^2$.
Orthogonality to the common regression space gives
\[
 \widehat r_{ij}-\frac1n\sum_k
 \varepsilon_i^{(k)}\varepsilon_j^{(k)}
 =-\frac1n(\varepsilon_i-e_i)^\top(\varepsilon_j-e_j),
\]
whose absolute value is at most $C_Mp\log(np)/n$.
Expanding the product difference and using the projection bounds
also gives
\begin{align*}
 \left\{\frac1n\sum_k
 (e_i^{(k)}e_j^{(k)}
  -\varepsilon_i^{(k)}\varepsilon_j^{(k)})^2\right\}^{1/2}
 &\le C_M\left\{
 \frac{\sqrt p\log(np)}{\sqrt n}
 +\frac{p^{3/2}\{\log(np)\}^{3/2}}n\right\}\\
 &\le C_M\frac{\sqrt p\log(np)}{\sqrt n}.
\end{align*}
This bounds the difference
between the two empirical standard deviations. Moreover, the population-product empirical variance is $\theta_{ij}\{1+O(\sqrt{\log(np)/n})\}$
on $A_1$. Therefore,
\[
 \left|\frac{\wtheta_{ij}}{\theta_{ij}}-1\right|
 \le C_M\frac{\sqrt p\log(np)}{\sqrt n}.
\]
These comparisons hold on the same event for every eligible
$U(i,j)$.

Define $\widehat Z_{ij}$ by replacing $e_i,e_j$ with
$\varepsilon_i,\varepsilon_j$ in $\widehat T_{ij}$. Whenever its denominator is positive,
\[
 |\widehat Z_{ij}|\ge t
 \quad\Longleftrightarrow\quad
 \frac{\left|\sum_k\varepsilon_i^{(k)}
                       \varepsilon_j^{(k)}\right|}
 {\left\{\sum_k(\varepsilon_i^{(k)}
               \varepsilon_j^{(k)})^2\right\}^{1/2}}
 \ge\frac{t}{\sqrt{1+t^2/n}}.
\]
For each null pair, the standardized population products are i.i.d
across samples, with mean zero, variance one, and uniformly bounded
third absolute moments. The self-normalized moderate-deviation theorem of \citet{jing2003self} gives a normal-tail approximation for this ratio with relative error $O((1+t)^3/\sqrt n)$. The threshold transformation changes the normal tail by relative $O((1+t)^4/n)$. Both errors vanish uniformly for $0\le t\le(M+1)\sqrt{\log(np)}$. The zero-denominator event is contained in $A_1^c$ for large $n$,
and
\[
 \frac{\mathbf{Pr}(A_1^c)}
 {1-\Phi((M+1)\sqrt{\log(np)})}
 \le C_M\sqrt{\log(np)}\,
 (np)^{-(M+1)^2/2-4}\longrightarrow0.
\]
It follows that, uniformly over null pairs and this range of $t$,
\[
 \mathbf{Pr}(|\widehat Z_{ij}|\ge t)
 =2\{1-\Phi(t)\}\{1+o(1)\}.
\]

Finally, both empirical variances are bounded away from zero on
$A_1$. The numerator and variance comparisons above imply
\[
 |\widehat T_{ij}-\widehat Z_{ij}|
 \le\delta_n(1+|\widehat Z_{ij}|),
 \qquad \delta_n=C_M\frac{p\log(np)}{\sqrt n}.
\]
Since $G\subseteq\widehat G$, this applies to the conditioning
sets used by the algorithm. For sufficiently large $n$,
\begin{align*}
 &\mathbf{Pr}\!\left(|\widehat Z_{ij}|\ge
 \frac{t+\delta_n}{1-\delta_n}\right)-\mathbf{Pr}(A_1^c)\\
 &\quad\le\mathbf{Pr}(|\widehat T_{ij}|\ge t)\\
 &\quad\le\mathbf{Pr}\!\left(|\widehat Z_{ij}|\ge
 \frac{\max\{t-\delta_n,0\}}{1+\delta_n}\right)
 +\mathbf{Pr}(A_1^c).
\end{align*}
The logarithmic derivative of the normal upper tail is
$O(1+t)$, and
$\delta_n\{1+M\sqrt{\log(np)}\}^2\to0$.
Thus both shifted normal tails are uniformly equivalent to
$2\{1-\Phi(t)\}$ for $0\le t\le M\sqrt{\log(np)}$.
The exceptional probability is negligible by the preceding bound.
Dividing the sandwich by $2\{1-\Phi(t)\}$ proves the proposition.
\end{proof}

\subsection{Proof of Lemma \ref{lem:rhoij}}
\label{sec:rhoij}
\rhoij*
\begin{proof}
The bounds are homogeneous under common rescaling, so it suffices
to prove them with $\svmin(\Sigma)=1$ and
$\svmax(\Sigma)\le C_2$. As in Section~\ref{sec:Tstat},
the global asymptotic convention implies
\[
 p^2\log(np)=O(n^{4/5}),
 \qquad
 \frac{p\{\log(np)\}^2}{\sqrt n}\longrightarrow0.
\]
Throughout the proof, $c,C>0$ depend only on the common
distributional bounds.

Let $\varepsilon_i$ be the population error in
Assumption~\ref{asump:indep}. The local Markov property and this
assumption give
$\varepsilon_i\perp X_{[p]\setminus\{i\}}$, as established in
Section~\ref{sec:Tstat}.
The full nodewise regression coefficients are
$-\frac{(\Sigma^{-1})_{i,-i}}{(\Sigma^{-1})_{ii}}$, and vanish outside the true neighborhood.
The block normal equations give
\begin{align*}
 \begin{pmatrix}e_i^0\\e_j^0\end{pmatrix}
 &=\big[(\Sigma^{-1})_{\{i,j\},\{i,j\}}\big]^{-1}
 \begin{pmatrix}
 (\Sigma^{-1})_{ii}\varepsilon_i\\
 (\Sigma^{-1})_{jj}\varepsilon_j
 \end{pmatrix},\\
 \Cov\!\begin{pmatrix}e_i^0\\e_j^0\end{pmatrix}
 &=\big[(\Sigma^{-1})_{\{i,j\},\{i,j\}}\big]^{-1}.
\end{align*}
The corresponding regression coefficients vanish outside
$(\operatorname{ne}(i)\cup\operatorname{ne}(j))\setminus\{i,j\}$.
Thus these are also the population residuals for every conditioning
set in the lemma. The covariance identity gives the stated
population correlation and variance bounds.

The coefficient matrix multiplying $(\varepsilon_i,\varepsilon_j)$
has operator norm at most $\kappa(\Sigma)\le C_2$.
The sub-Gaussian bounds for the nodewise errors established in
Section~\ref{sec:Tstat} therefore also hold uniformly for
$e_i^0,e_j^0$.
Their sample means, covariate scores, and maxima are bounded linear
combinations of the corresponding node-error quantities.
Consequently, the covariance, sample-mean, score, and maximum
bounds used in Section~\ref{sec:Tstat} imply, outside an event
of probability at most $C(np)^{-2}$,
\begin{align*}
 \frac{\|e_i^0-e_i\|_2}{\sqrt n}
 &\le C\sqrt{\frac{p\log(np)}n},\\
 \max_k|(e_i^0)^{(k)}-e_i^{(k)}|
 &\le C\frac{p\log(np)}{\sqrt n},\\
 \max_k|(e_i^0)^{(k)}|
 &\le C\sqrt{\log(np)},
\end{align*}
and the same bounds hold for $j$, simultaneously over all pairs
and eligible conditioning sets.
Here population residuals in vector expressions denote their
sampled vectors. For completeness, the first two bounds follow
from the normal equations: the score vector has Euclidean norm
at most $C\sqrt{p\log(np)/n}$, the inverse Gram matrix has
operator norm at most $2$, and each design row has Euclidean norm
at most $C\sqrt{p\log(np)}$.
The full Gram bound applies to all principal submatrices; the
sample-mean bounds give the same result when a constant column
is included. Thus the argument covers either intercept convention.

Orthogonality to the common regression space gives
\[
 \widehat r_{ij}
 -\frac1n\sum_{k=1}^n(e_i^0)^{(k)}(e_j^0)^{(k)}
 =-\frac1n(e_i^0-e_i)^\top(e_j^0-e_j).
\]
Its absolute value is at most $Cp\log(np)/n$.
The population pair products are sub-exponential, so Bernstein's
inequality and a union bound over pairs give
\[
 \max_{i\neq j}
 \left|\frac1n\sum_{k=1}^n(e_i^0)^{(k)}(e_j^0)^{(k)}
 -\E[e_i^0e_j^0]\right|
 \le C\sqrt{\frac{\log(np)}n}
\]
outside an event of probability at most $C(np)^{-2}$.
This proves the claimed bound for $\widehat r_{ij}$.

To control the denominator, the nodewise regression equation gives
\[
 e_i^0=-\frac{(\Sigma^{-1})_{ij}}{(\Sigma^{-1})_{ii}}
 e_j^0+\varepsilon_i,
 \qquad \varepsilon_i\perp e_j^0,
\]
because $e_j^0$ is a function of $X_{[p]\setminus\{i\}}$.
Since $\varepsilon_i$ is centered,
\begin{align*}
 \Var(e_i^0e_j^0)
 &=\frac{(\Sigma^{-1})_{ij}^2}{(\Sigma^{-1})_{ii}^2}
 \Var((e_j^0)^2)
 +\Var(\varepsilon_i)\Var(e_j^0)\\
 &\ge1.
\end{align*}
The sub-Gaussian moment bounds also give
$\Var(e_i^0e_j^0)\le C$ and $\E|e_i^0e_j^0|^4\le C$.
Chebyshev's inequality for the empirical first and second product
moments, followed by a union bound over pairs, therefore shows
that all centered empirical variances of the population products
lie between two positive constants, outside an event of
probability at most $Cp^2/n$.

Expanding the fitted-product difference and using the preceding
projection and maximum bounds yields
\begin{align*}
 \left\{\frac1n\sum_{k=1}^n
 \bigl(e_i^{(k)}e_j^{(k)}
 -(e_i^0)^{(k)}(e_j^0)^{(k)}\bigr)^2\right\}^{1/2}
 &\le C\left\{
 \frac{\sqrt p\log(np)}{\sqrt n}
 +\frac{p^{3/2}\{\log(np)\}^{3/2}}n\right\}
 \longrightarrow0,
\end{align*}
uniformly over all eligible conditioning sets.
Centering contracts the Euclidean norm, so this also bounds the
difference between the fitted-product and population-product
empirical standard deviations. Hence
$c\le\sqrt{\wtheta_{ij}}\le C$ simultaneously on the same event
for sufficiently large $n$.
Combining the exceptional probabilities and using
$(np)^{-2}\le p^2/n$ gives total failure probability at most
$Cp^2/n$. Rescaling back completes the lemma.
\end{proof}

\section{Experiment details}\label{Expt details}
\subsection{Set-up}\label{Expt setup}
\paragraph{Computing}
All experiments were conducted on an 16-core Intel Xeon Gold 6248R processor with
3.00 GHz base frequency and 16GB of memory. All experiments had a 36 hour wall clock limit. 

\paragraph{Graph Models}
We used four common graph ensembles: Band, Scale-Free (SF), Erdös-Rényi (ER) and k-Nearest Neighbor (KNN) graphs.
The Band and SF graph structure follow the flare~\citep[][]{flare} package, and the ER and KNN graphs follow i-graph~\citep[][]{igraph} and mstknnclust~\citep[][]{knn} in R. Specifically, graphs are initialized as following
\begin{itemize}
    \item Band graphs: These are single chain graph given the number of observations $n$ and the number of variables $p$, and $u,v$ are set as default in the r-flare package and bandwidth is set to be 1.
    \item Scale-free (SF) graphs: These are generated by Barabási–Albert model; The graph is initialized with two connected nodes, and the probability of a new node connecting to one existing node is proportional to the degree of the existing node.
    \item Erdös-Rény (ER) graphs: These are random undirected graph with $p$ number of nodes, and $\lceil\frac{p\log p}{10}\rceil$ number of edges, which are selected uniformly and randomly from the set of possible edges. 
    \item k-nearest neighbor (KNN) graphs: There are random graphs where nodes are connected only if they are one of the $k$-nearest neighbors based on corresponding distance between them. A uniformly distributed $p\times p$ matrix is generated as some random data to calculate euclidean distances between nodes. We set $k=1$ to construct sparse graphs.
\end{itemize}

\paragraph{Method}
We consider the following methods for $\lambda$ tuning.
\begin{itemize}
    \item Leave-one-out cross-validation (LOOCV): Each single data point is removed to form the validation set, and the rest are used for training set.
    \item $K$-fold cross-validation (CV $K$-fold): We set $K=3,5,10$, and each fold is used for validation set, and the rest folds are for training set. We use 3 folds as default and denote it by CV $K$-fold, and use CV\_3fold in particular when comparing with other number of folds.
    \item Approximate cross-validation (ACV): We consider ACV, ProxACV~\citep[][]{wilson2020approximate} and ACV $K$-fold which serves as a heuristic extension of ACV for large $n$ and $p$.
    \item (extended) Bayesian information criteria (BIC/EBIC): The log-likelihood is estimated by $\lambda=0$ with zero support constraint in R package~\citep[glasso][]{glasso}, and $\gamma = 1$ for EBIC based on equation 1 in~\citep[][]{foygel2010extended}.
\end{itemize}

\paragraph{Remarks}
We use the glasso R package~\cite{glasso} to estimate the precision matrix, and the code is available at:~\url{https://github.com/cran/glasso}. We use QUIC~\citep[][]{hsieh2014quic} for ProxACV following the same construction in~\citep[][]{wilson2020approximate}. The set of $\lambda$ values is pre-computed. The $\lambda_{\max}$ is the maximum value of the infinity norm of the correlation between each node and the rest~\citep[][]{friedman2008sparse}. We decrease $\lambda_{\max}$ with a fixed ratio to find $\lambda_{\min}$ that results in a trivial full graph estimate by glasso. Time is plotted in log-scale. 

We note that for each candidate support, BIC/EBIC requires computing the maximum likelihood solution through an additional support-constrained unpenalized GLasso optimization. This necessitates an additional cubic cost from computing the Gaussian likelihood. In contrast, ACV constructs each approximate fold estimate using the explicit one-step update in \eqref{eq:acv}, without another optimization or convergence loop. Hence, we expect a computational advantage which comes from replacing additional iterative constrained solves in GLasso with a fixed number of explicit updates.

\subsection{Sub-Gaussian simulation}
\label{sec:appendix subG}

\begin{figure}[h!]
  \centering
  \includegraphics[width=\textwidth]{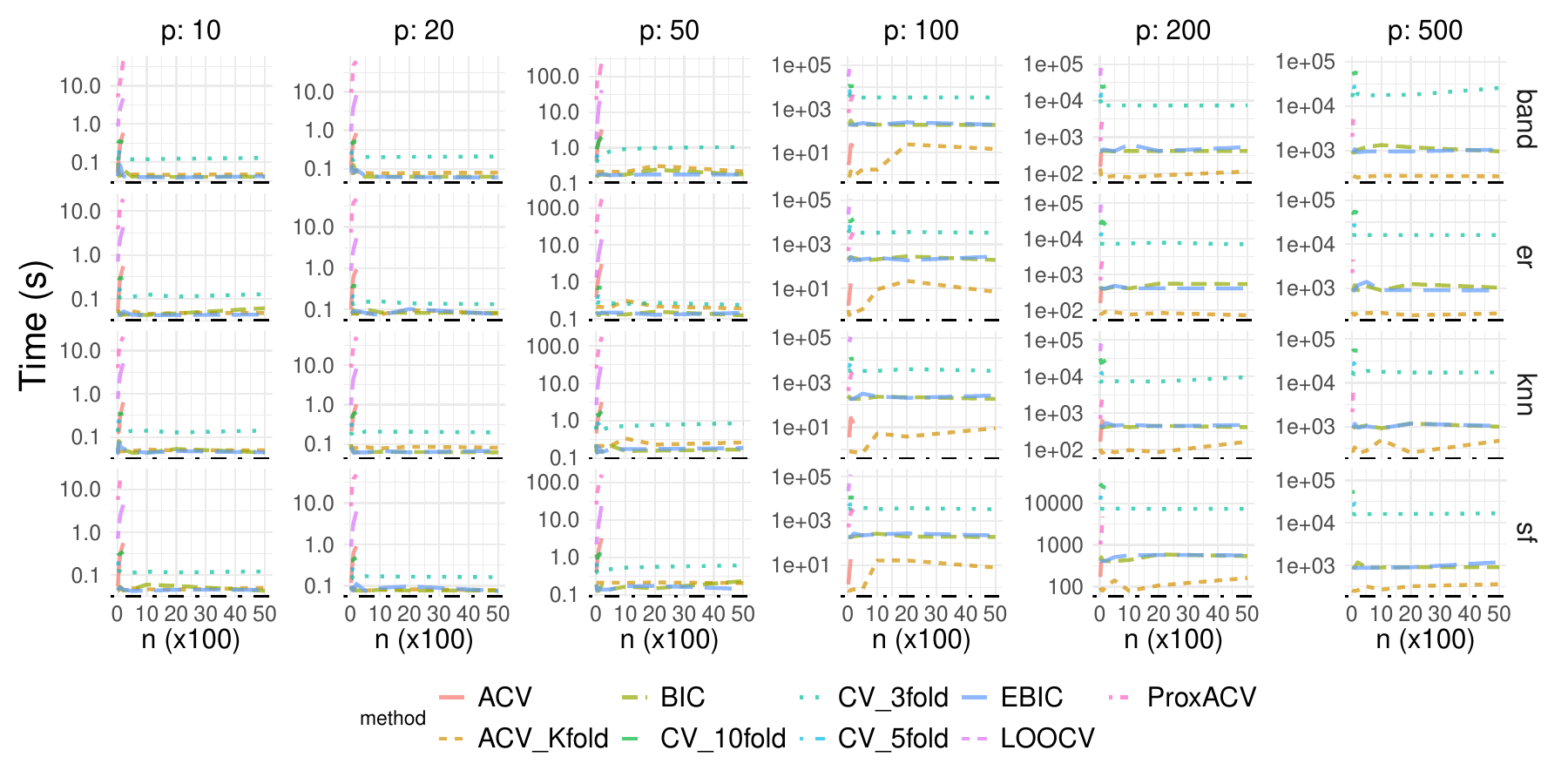}
  \caption{Evaluations of methods via computational cost in large scale ($n$ up to 5000) under sub-Gaussian distribution. LOOCV, ACV and ProxACV have computational expenses scaling with $n$ and quickly become infeasible. ACV $K$-fold has constantly much less computational cost compared to CV $K$-fold and BIC/EBIC.}
  \label{fig: EXP2 laplace time large}
\end{figure}
We consider the Gaussian distribution for sub-Gaussian case. Specifically, we follow flare package \citep[][]{flare} to construct Gaussian distributed data with mean 0 and covariance matrix based on the graph structure. We set $n$ from 10 to 5000, and $p = (10,20,50,100,200,500)$. 

\begin{figure}[!h]
  \centering
  \includegraphics[width=\textwidth]{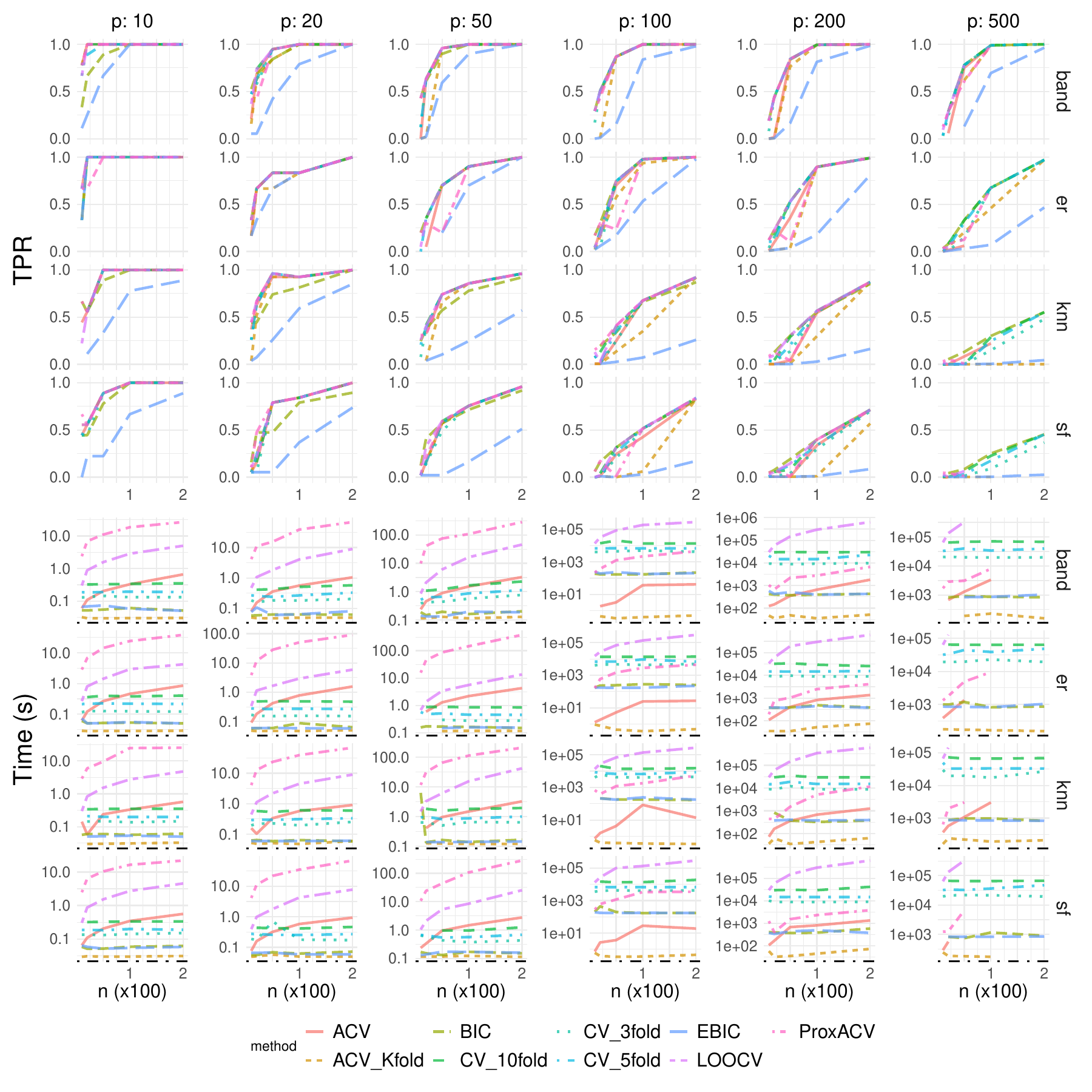}
  \caption{Comparison between LOOCV variants (ACV and ProxACV) and other methods under sub-Gaussian distribution. While LOOCV, ACV and ProxACV have best performance in discerning edges based on TPR, their computational expenses scale with $n$.}
  \label{fig: EXP2 laplace small}
\end{figure}

To compare all CV variants in a computationally feasible set, we constraint $n\leq 200$ to show pros and cons of LOOCV and its variants, ACV and ProxACV in Figure~\ref{fig: EXP2 laplace small}. Clearly, LOOCV and its variants have the fastest convergence to TPR = 1 compared to $K$-fold CV, BIC and EBIC. But their computational costs all scale with $n$, quickly becoming impractical for $n\geq 100$. In contrast, BIC and EBIC requires much more data to reach a TPR level similar to LOOCV and its variants. However, their computational costs are much more affordable.

\begin{figure}[h]
  \centering
  \includegraphics[width=0.85\textwidth]{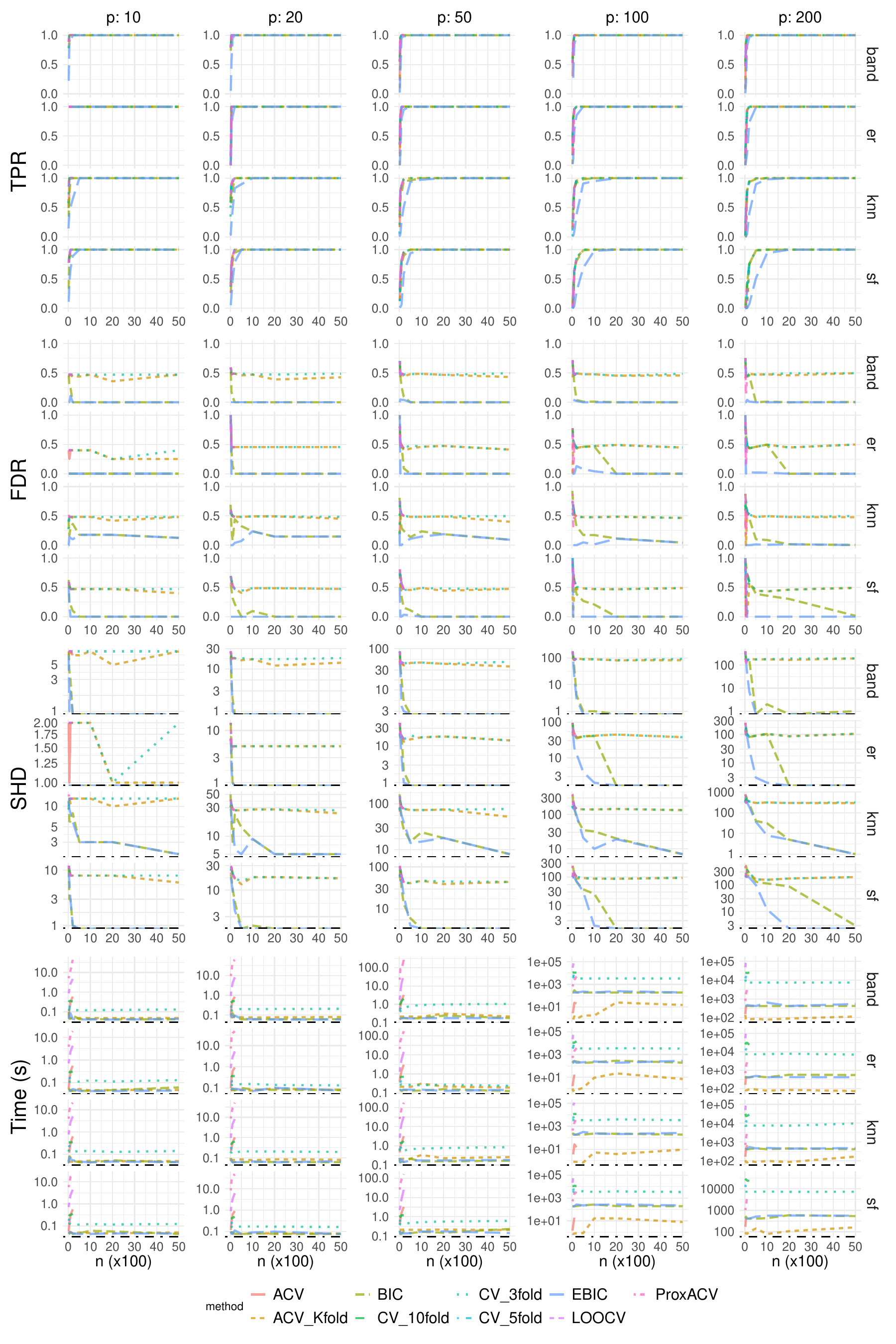}
  \caption{Evaluations of methods via performance in large scale ($n$ up to 5000) under sub-Gaussian distribution. Average SHD, TPR and FDR over varying sample size $n$ (in hundreds) and the number of nodes $p$ on groups of graph types. The black dotdash line represents the $0$-SHD, i.e. perfect structure learning. All CV variants and approximations alone are inconsistent with non-zero FDR and SHD.}
  \label{fig: EXP2 laplace large}
\end{figure}

We show the comprehensive results for $n$ up to 5000 to compare ACV extended to $K$-fold CV (ACV $K$-fold) concerning performance and computational cost in Figure~\ref{fig: EXP2 laplace large} and Figure~\ref{fig: EXP2 laplace time large} respectively. The main message of Figure~\ref{fig: EXP2 laplace time large} is that ACV $K$-fold has least computational cost. Moreover, Figure~\ref{fig: EXP2 laplace large} shows that CV variants and CV approximations are successful in pre-screening, but call out for a post-correction because they are inconsistent regardless of the growing sample size.

\begin{figure}[h]
  \centering
  \includegraphics[width=\textwidth]{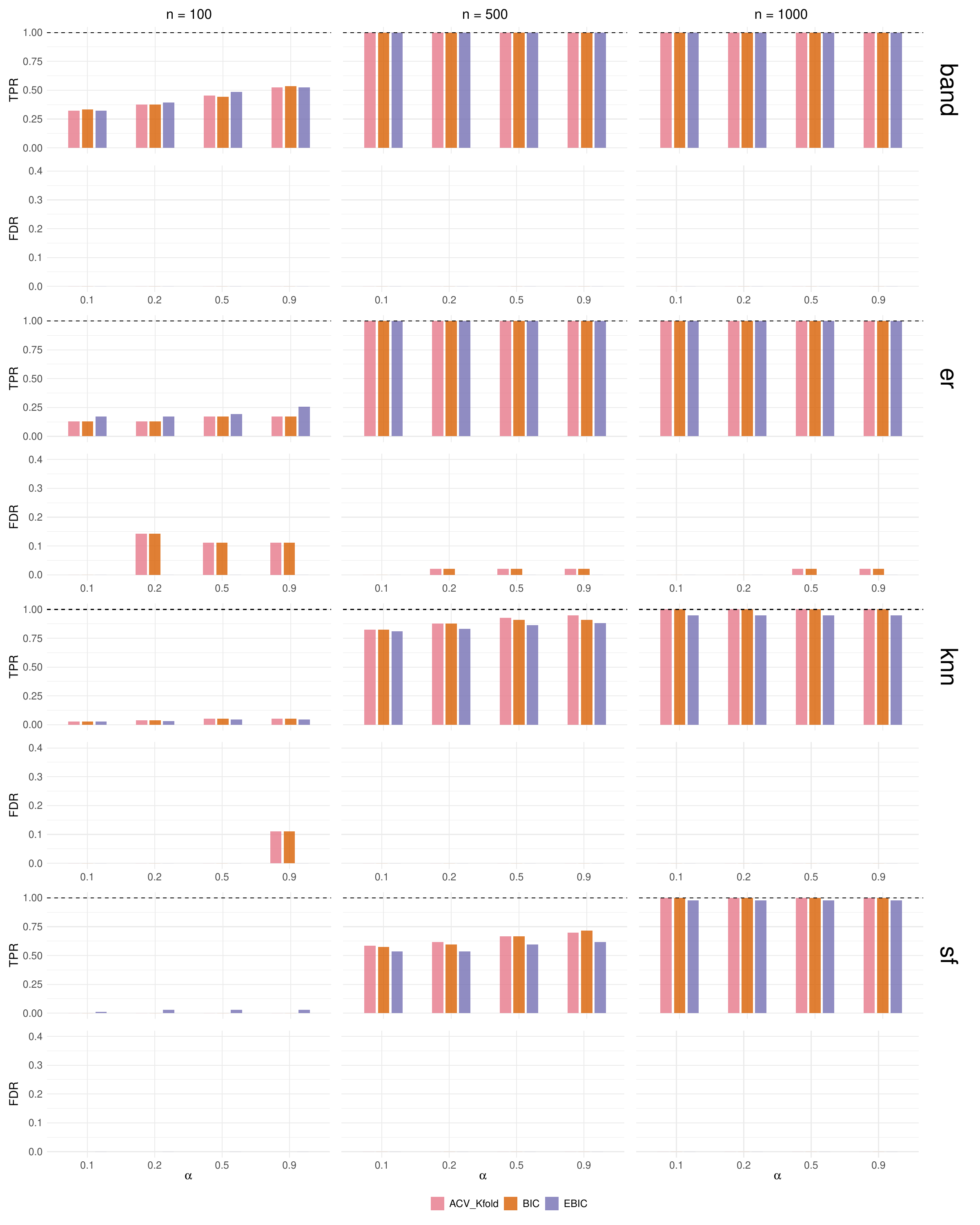}
  \caption{Performance comparison between ACV $K$-fold, BIC and EBIC incorporated with the post-correction with varying $n$ and fixed $p=100$ under  sub-Gaussian distribution. $\alpha$ level is pre-set and dashed black line is TPR = 1.}
  \label{fig: EXP3 laplace fdr}
\end{figure}

The post-correction with pre-set $\alpha$ level at fixed $p=100$ is shown in Figure~\ref{fig: EXP3 laplace fdr}. $n$ choices are made to summarize the super-set relations between the graph estimate and the true graph. However, we note that different graph ensembles encode varying learning difficulties, and we roughly set $n=\{100,500,1000\}$ to represent the three cases of the graph estimate $\wG$:
\begin{enumerate}
    \item not super-set at $n=100$: $\wG\not\supseteq\G$.
    \item roughly a super-set at $n=500$:  $\wG\mathrel{\substack{
  \textstyle\supset\\[-0.2ex]\textstyle\sim}}\G$.
    \item a super-set at $n=1000$: $\wG\supseteq\G$.
\end{enumerate}As stated in Section~\ref{sec:fdr control}, the family-wise false rate can be loose, and it is illustrated in Figure~\ref{fig: EXP3 laplace fdr} that the FDR level is way below the $\alpha$ thresholding. Three sample choices ($n=100, 500, 1000$) correspond to three different super set relations between the graph estimate and the true graph. When $\G\not\subseteq\wG$ ($n=100$), the post-correction screens harshly and substantially hurts TPR. In contrast, when $\wG$ roughly contains the true graph $\G$, the post-correction effectively removes false discoveries while remaining a reasonable TPR. By comparison, we see that ACV $K$-fold incorporated with the post-correction outperforms BIC and EBIC in both computational cost and structure learning performance. For example, when $\alpha = 0.5$, ACV $K$-fold has greater TPR compared to BIC and EBIC while obtaining 0 FDR.

\clearpage
\subsection{Non sub-Gaussian simulation}
\label{sec:appendix lognormal}
We detail the log-normal distribution case for robustness testing to the sub-Gaussian assumption. We use the stats package in R to construct log-normal distributed data with mean 0 and standard deviation $1$. We follow the same $n,p$ setting in the sub-Gaussian case for comparison.

\begin{figure}[!h]
  \centering
  \includegraphics[width=\textwidth]{{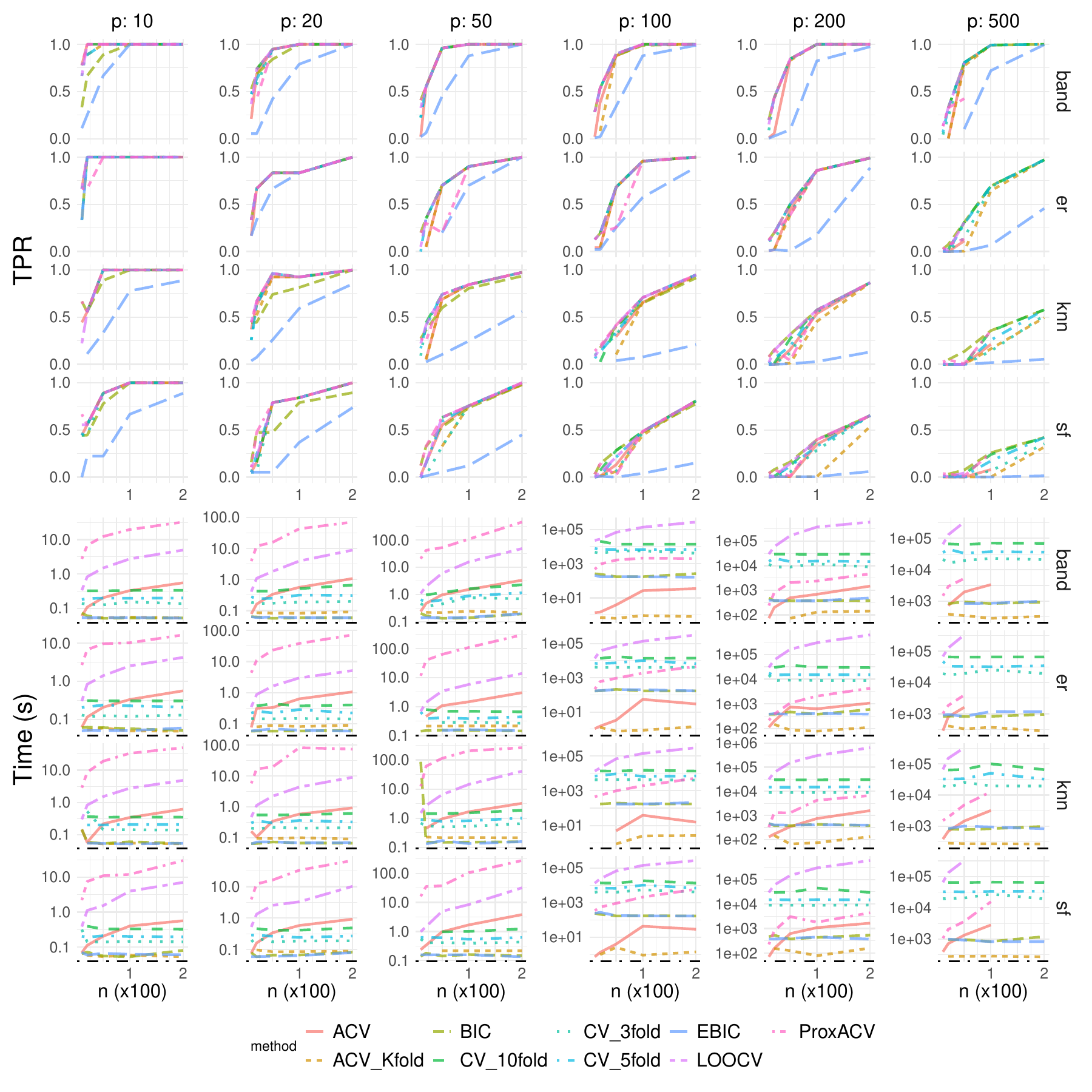}}
  \caption{Comparison between LOOCV variants (ACV and ProxACV) and other methods under log-normal distribution. While LOOCV, ACV and ProxACV have best performance in discerning edges based on TPR performance, their computational expenses scale with $n$. Time is plotted in log-scale.}
  \label{fig: EXP2 lognormal small}
\end{figure}

\begin{figure}[h]
  \centering
  \includegraphics[width=\textwidth]{{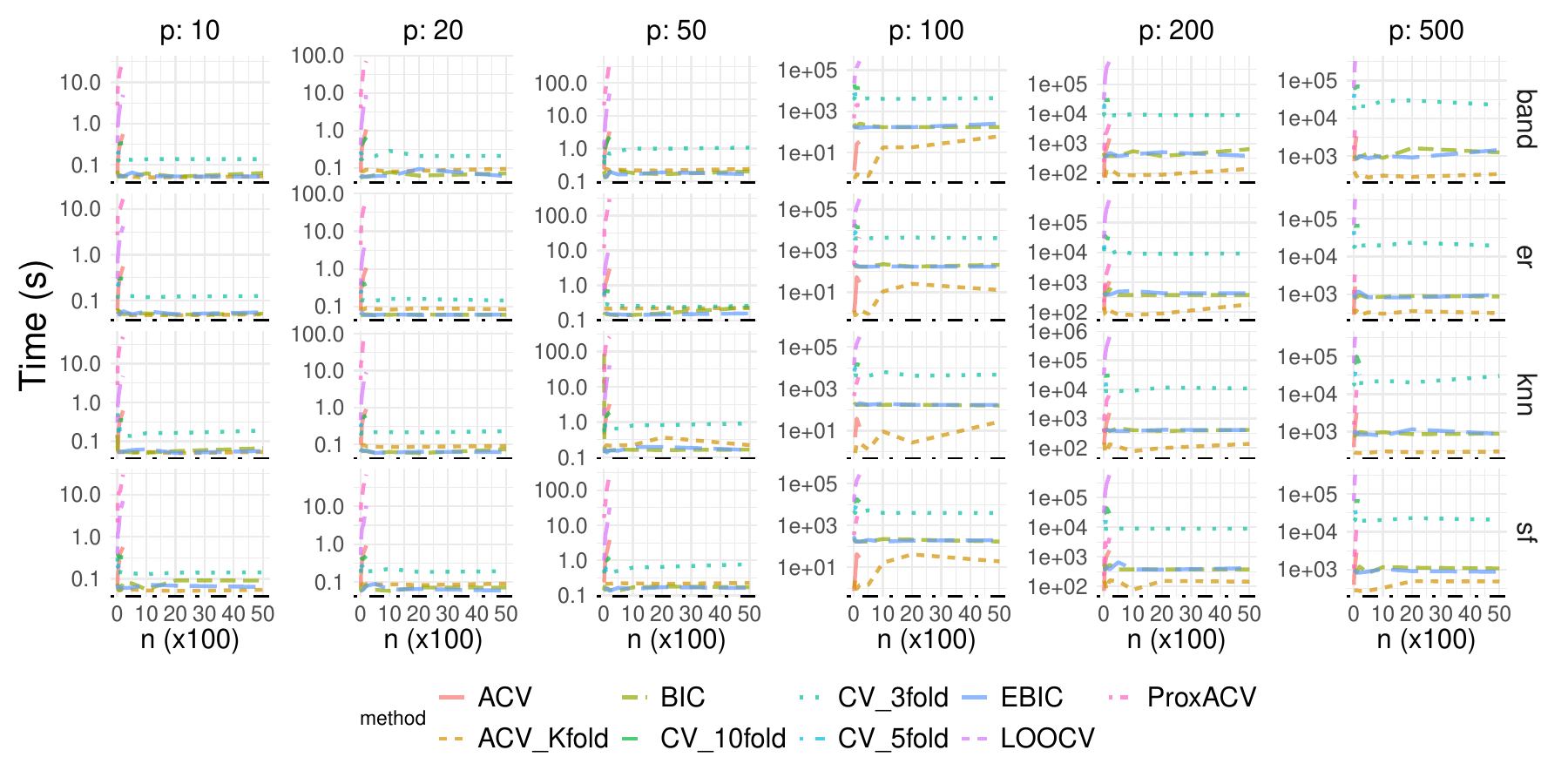}}
  \caption{Evaluations of methods via computational cost in large scale ($n$ up to 5000) under log-normal distribution. LOOCV, ACV and ProxACV have computational expenses scaling with $n$ and quickly become infeasible. ACV $K$-fold has constantly much less computational cost compared to $K$-fold CV and BIC/EBIC. Time is plotted in log-scale.}
  \label{fig: EXP2 lognormal time large}
\end{figure}

\begin{figure}[!h]
  \centering
  \includegraphics[width=0.85\textwidth]{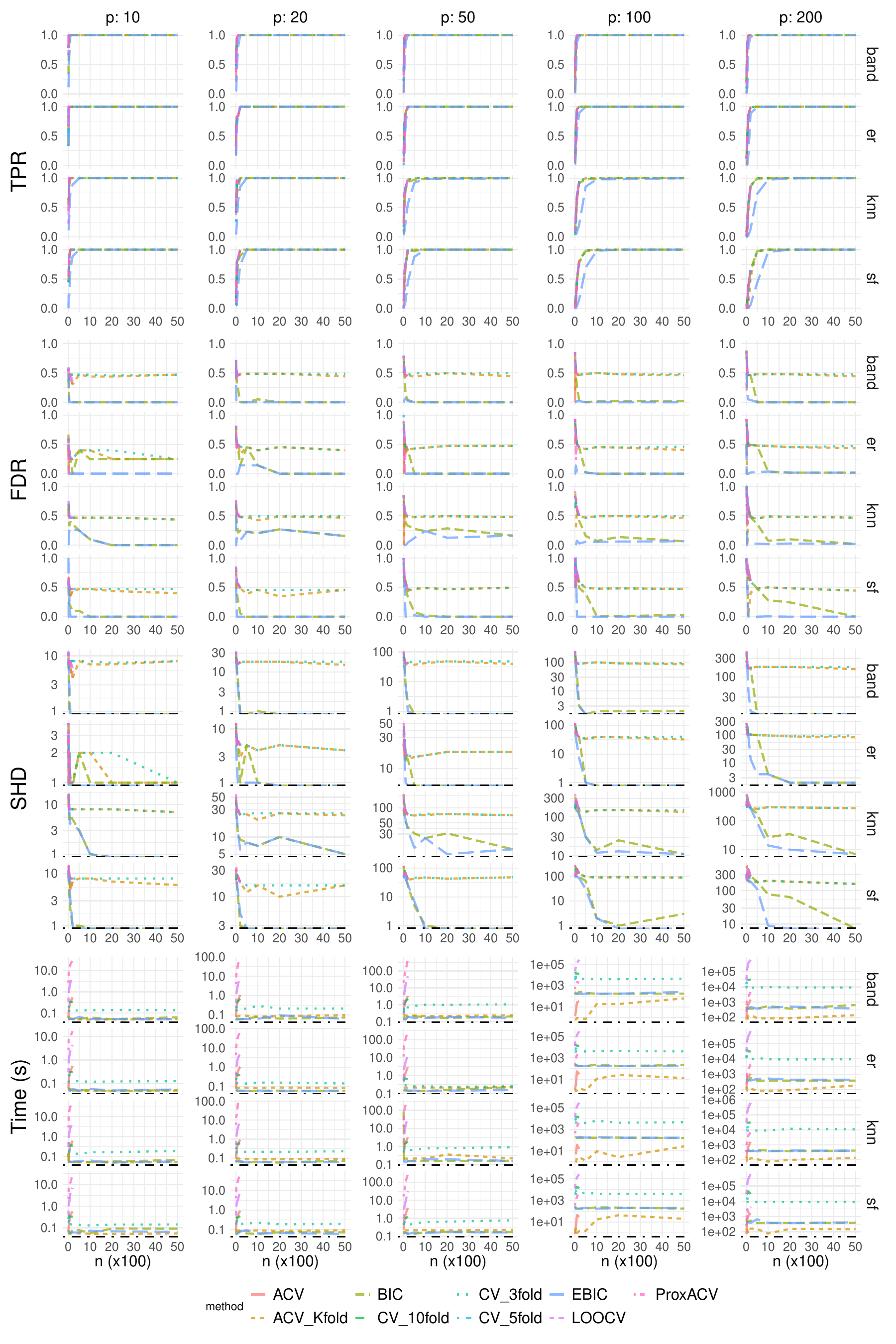}
  \caption{Evaluations of methods via performance in large scale ($n$ up to 5000) under log-normal distribution. Average SHD, TPR and FDR over varying sample size $n$ (in hundreds) and the number of variables $p$ on groups of graph types. The black dotdash line represents the $0$-SHD, i.e. perfect structure learning. All CV variants and approximations alone are inconsistent with non-zero FDR and SHD.}
  \label{fig: EXP2 lognormal large}
\end{figure}

Figure~\ref{fig: EXP2 lognormal small} shows the computational disadvantage of LOOCV, ACV and ProxACV while they obtain an outstanding performance in edge detection with less samples compared to $K$-fold CV, BIC and EBIC. We also show the computational cost and the structure learning performance for large sample size in Figure~\ref{fig: EXP2 lognormal time large} and Figure~\ref{fig: EXP2 lognormal large} respectively. The conveyed message is same as the sub-Gaussian case: ACV is a qualified and efficient pre-screener but requires the post-correction for consistency.

\begin{figure}[!h]
  \centering
  \includegraphics[width=\textwidth]{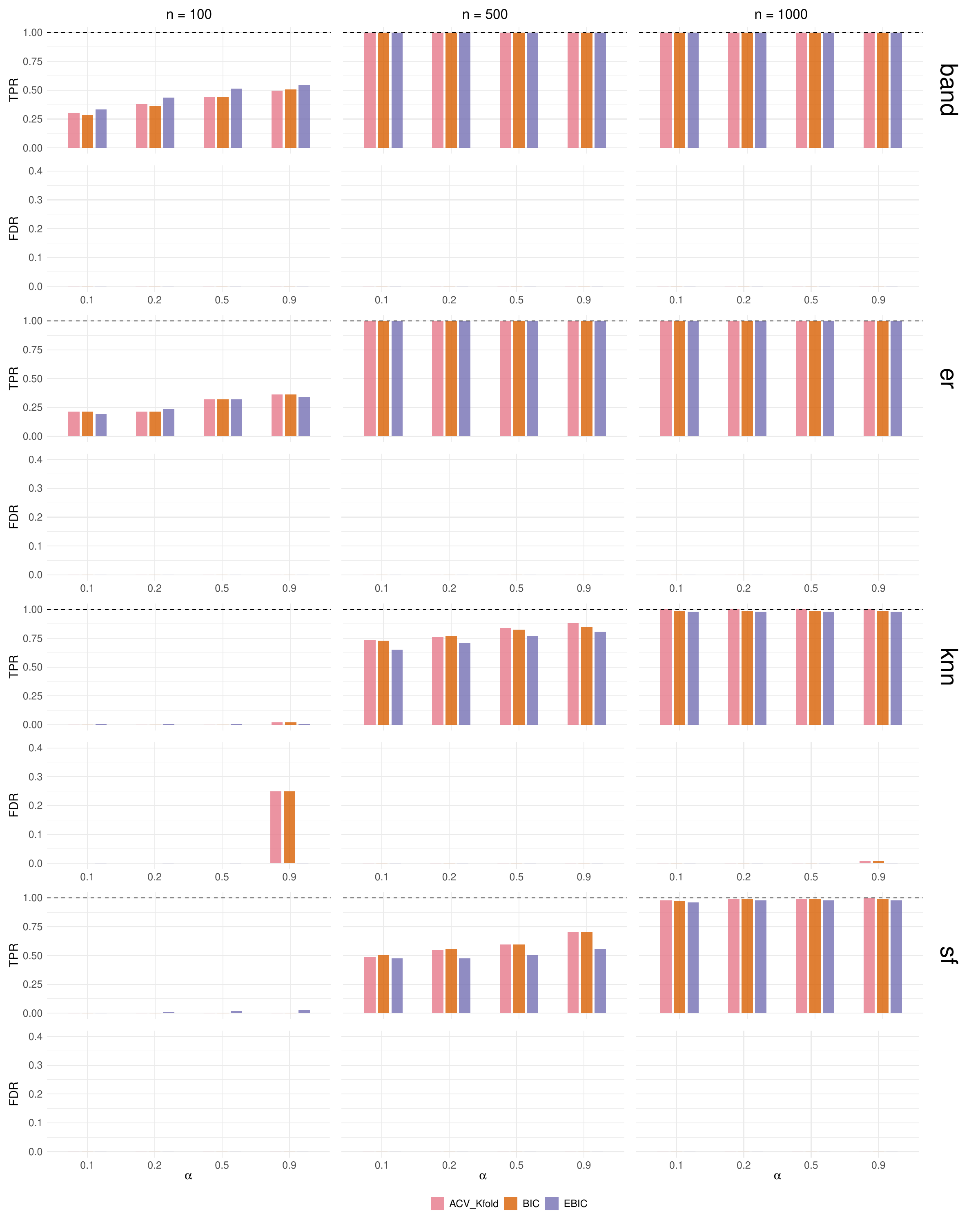}
  \caption{Performance comparison between ACV $K$-fold, BIC and EBIC incorporated with the post-correction with varying $n$ and fixed $p=100$ under log-normal distribution. $\alpha$ level is pre-set and dashed black line is TPR = 1.}
  \label{fig: EXP3 lognormal fdr}
\end{figure}

The evaluation of ACV $K$-fold, BIC and EBIC incorporated with the post-correction in illustrated in Figure~\ref{fig: EXP3 lognormal fdr}. Besides similar conclusion in the sub-Gaussian case, we observe that ACV $K$-fold discerns more true edges when data is insufficient for sparse graphs such as KNN and SF ($n=500$). This incredible ability inherited from CV for edge detection with limited samples is crucial to reach consistency. For example, when $\alpha = 0.5$, ACV $K$-fold is consistent for KNN and SF, while BIC and EBIC fail to uncover all true edges ($\text{TPR}\neq 1$).

\begin{figure}[hb]
  \centering
  \includegraphics[width=\textwidth]{{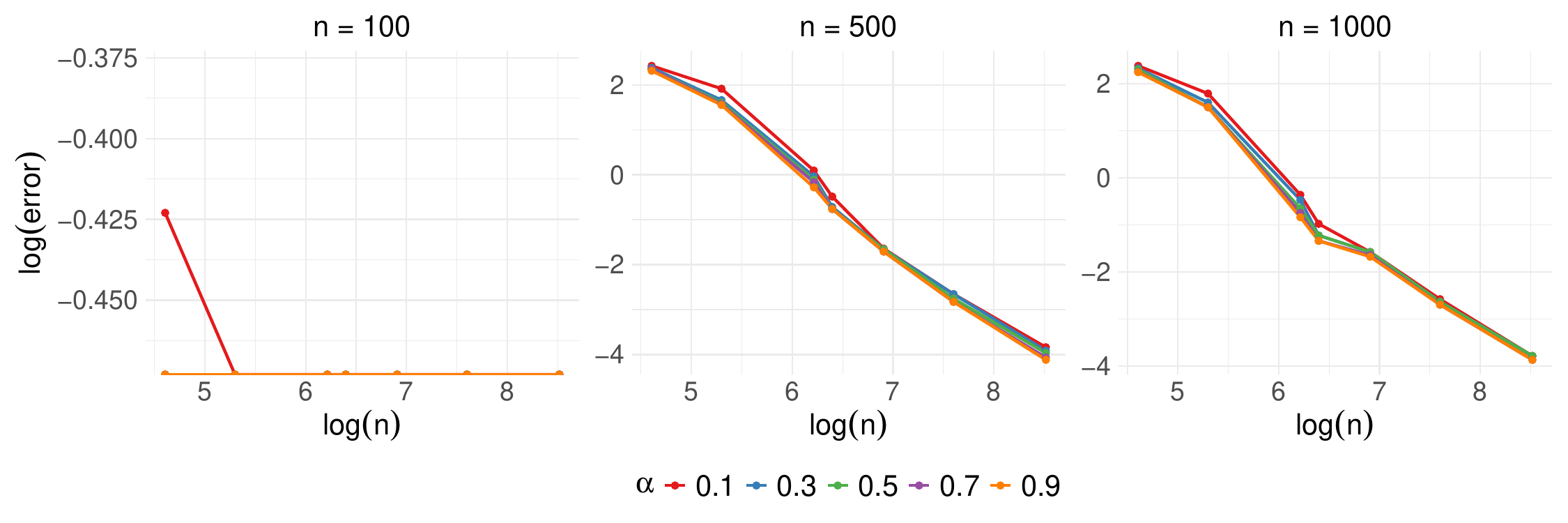}}
  \caption{Evaluation of $\alpha$ via predictive error depending on sample size $n$.}
  \label{fig: EXP4 alpha}
\end{figure}

\begin{figure}[!ht]
  \centering
  \includegraphics[width=\textwidth]{{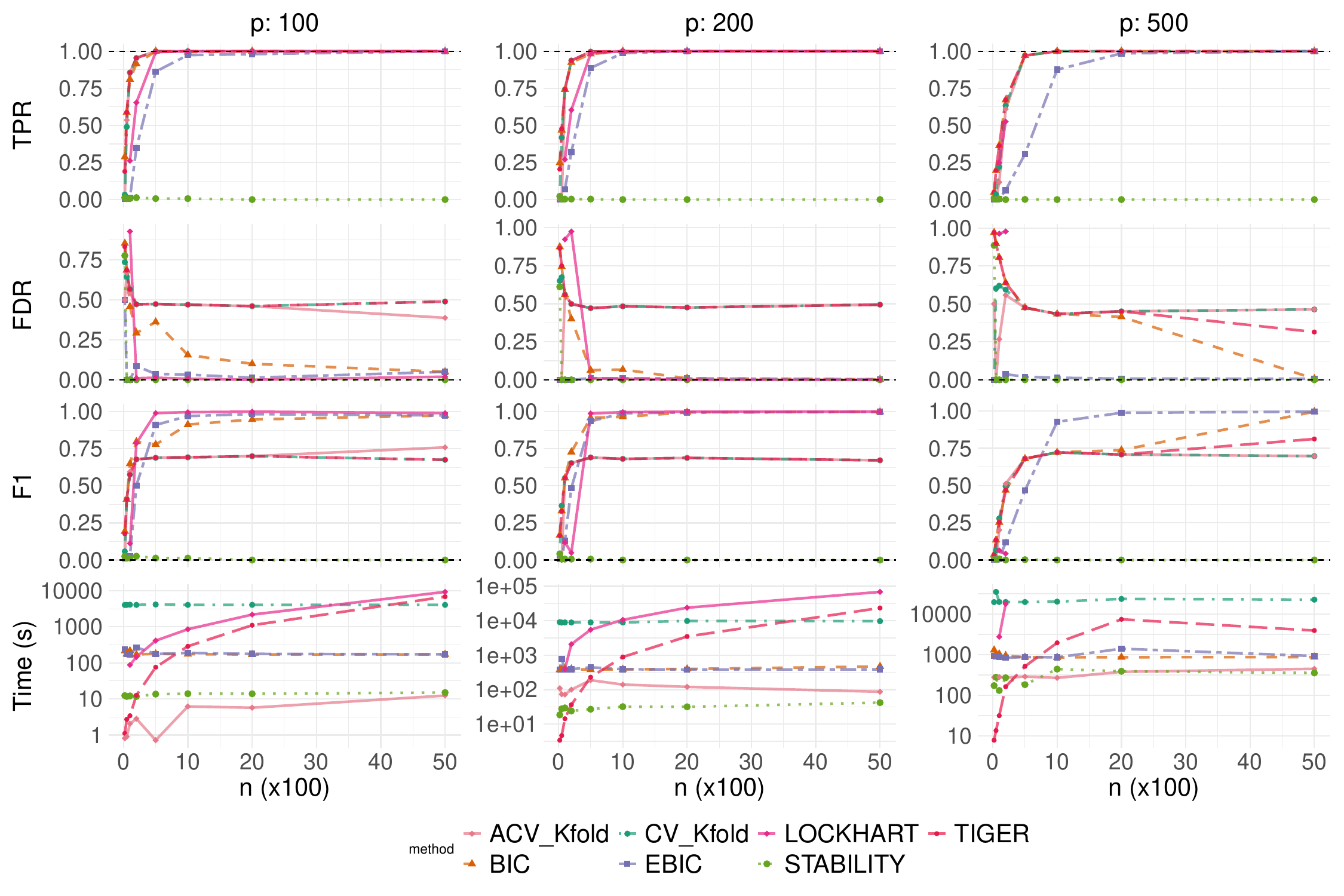}}
  \caption{Comparison of methods via performance on pre-screening under Gaussian distribution. ACV $K$-fold outperforms the rest in pre-screening concerning computational cost and samples needed.}
  \label{fig: EXP2 knn compare}
\end{figure}

\subsection{Non-linear simulation}
\label{sec:appendix ising}
We consider the Ising model for the nonlinear case. Each row of data $\bX\in\R^{n\times p}$ is i.i.d generated based on graph ensembles via the IsingSampler\citep[][]{IsingSampler}, and we construct $\ell_1$ penalized estimator via neighborhood-based logistic regression in \citep[][]{ravikumar2010high}. Let $\lambda > 0$. Without loss of generality, we estimate neighbors of $p$th node by regressing on others as the following:
\begin{align*}
    \wtheta^{\lambda,\bX}
    :&= \argmin_{\theta\in\R^{p-1}}\frac{1}{n}\sum_{i=1}^n f(\theta; \bX^{(i)}) + \lambda\norm{\theta}_1,\\
    f(\theta;\bX^{(i)})
    :&= \log\left(e^{\sum_{j=1}^{p-1} \theta_j\bX_j^{(i)}} + e^{-\sum_{j=1}^{p-1}\theta_j\bX_j^{(i)}}\right) -\sum_{j=1}^{p-1}\theta_j\bX_p^{(i)}\bX_j^{(i)}\\
    \wne^{\lambda,\bX}
    :&= \{j\in[p-1]\mid \wtheta^{\lambda,\bX}_j\neq 0\}.
\end{align*}
Once we have estimated neighbors $\wne^{\lambda,\bX}$ for each node, we can combine with the AND or the OR rule to symmetrize the estimated graph as discussed in \citep[][]{meinshausen2006high}. The derivation of the approximate CV is similar but to vector format, and detailed derivation can be found in Appendix D.4 in \citep[][]{stephenson2020approximate}.
\begin{align*}
    \Tilde{\theta}^{\lambda,\bX}
    :&= \begin{bmatrix}
        \wtheta^{\lambda,\bX}_{\hat{\mathcal{S}}} + (H^{(-i)}_{\hat{\mathcal{S}}\hat{\mathcal{S}}})^{-1}[\nabla_{\theta}f(\theta;\bX^{(i)})]_{\hat{}}\\
        \bm{0}_{\setminus\hat{\mathcal{\mathcal{S}}}}
    \end{bmatrix},
    \quad H^{(-i)}_{\hat{\mathcal{S}}\hat{\mathcal{S}}} = H_{\hat{\mathcal{S}}\hat{\mathcal{S}}} - [\nabla^2_{\theta}f(\theta;\bX^{(i)})]_{\hat{\mathcal{S}}\hat{\mathcal{S}}},
\end{align*}
where $H_{\hat{\mathcal{S}}\hat{\mathcal{S}}}$ is restricted Hessian on the support $\hat{\mathcal{S}} = \supp(\wtheta^{\lambda,\bX})$. 

\begin{figure}[!h]
  \centering
  \includegraphics[width=0.85\textwidth]{{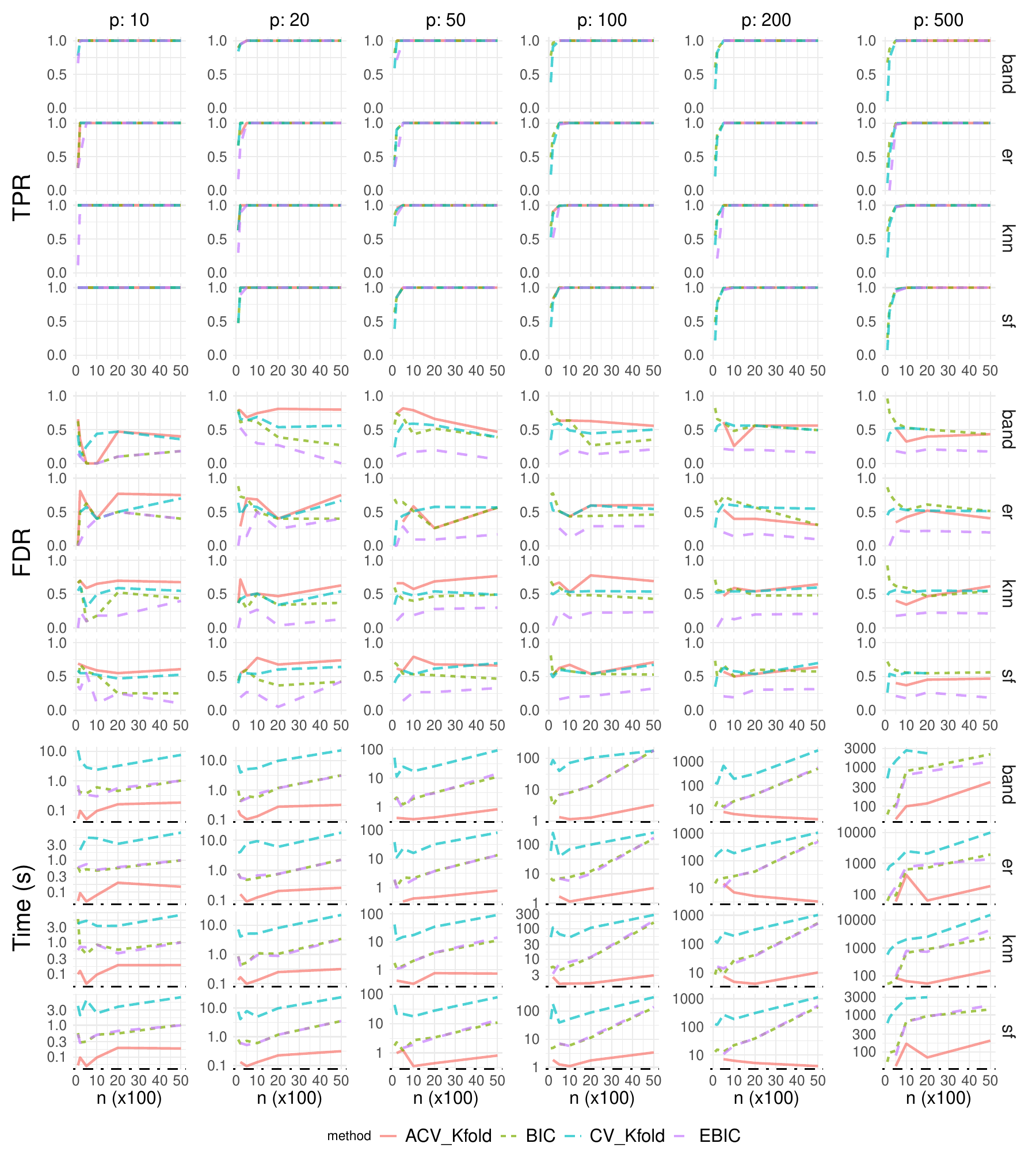}}
  \caption{Evaluations of methods via performance in large scale ($n$ up to 5000) under Ising model. Average TPR, FDR and Time in log-scale over varying sample size $n$ (in hundreds) and the number of variables $p$ on groups of graph types.}
  \label{fig: EXP2 nonlinear large}
\end{figure}

Different from the Kronecker product in \eqref{eq:acv}, the approximation involves complicated and expensive calculations for the Hessian inverse vector product. We implement preconditioned conjugate gradient method to approximate the Hessian inverse product for ACV $K$-fold, and the result is shown in Figure \ref{fig: EXP2 nonlinear large}. We note that this can be improved with other alternatives, and leave possible improvements for future work as it is orthogonal to our focus on robustness testing.

We find in Figure \ref{fig: EXP2 nonlinear large} that the computational improvement on Hessian inverse product calculation does not compromise the pre-screening ability, leaving ACV $K$-fold a qualified pre-screener for the post-correction. Moreover, Figure \ref{fig: EXP2 nonlinear large} shows that the computational advantage of ACV soon occurs and remains when both the number of observations and the number of variables growing large.

\begin{figure}[!h]
  \centering
  \includegraphics[width=\textwidth]{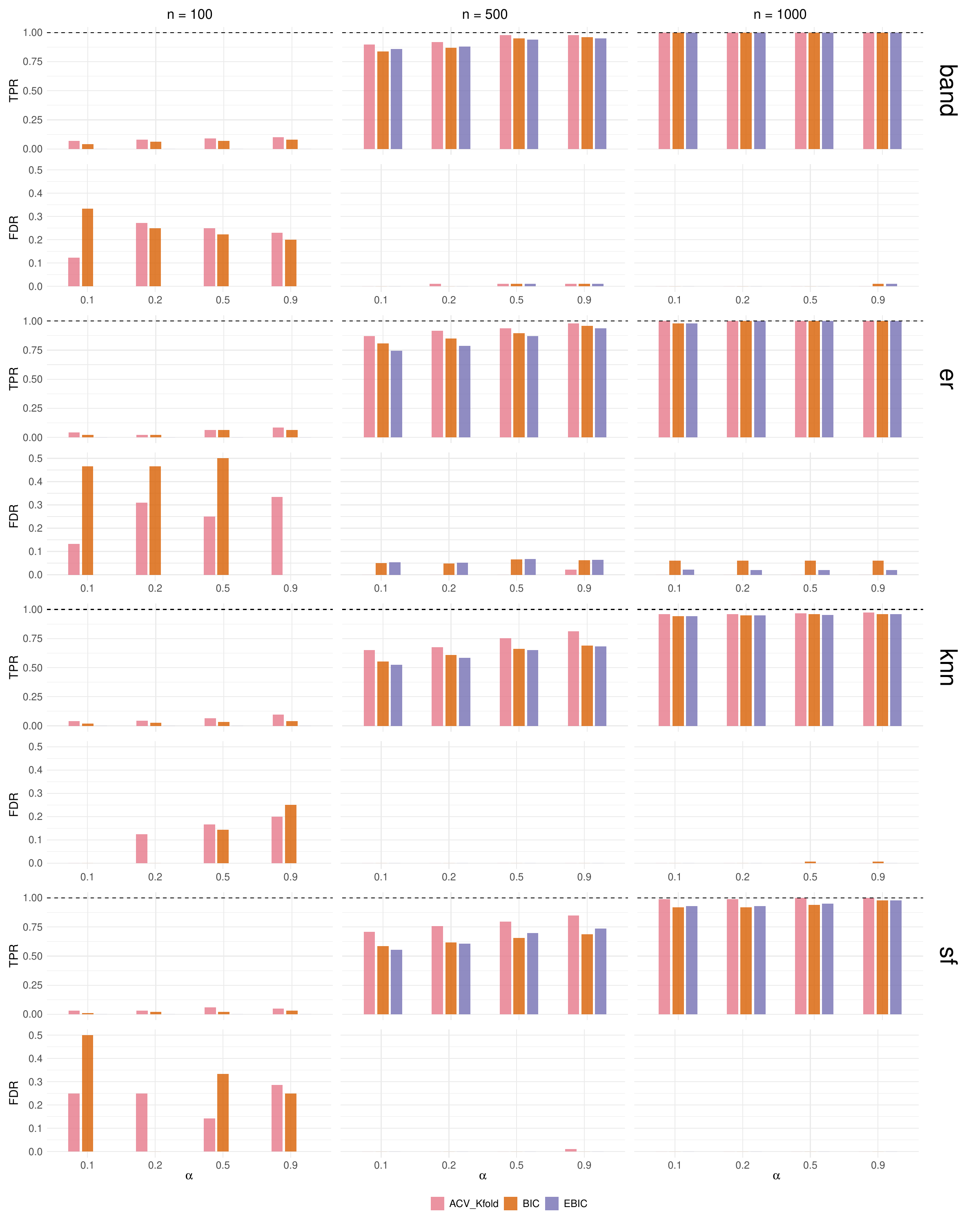}
  \caption{Performance comparison between ACV $K$-fold, BIC and EBIC incorporated with the post-correction with varying $n$ and fixed $p=100$ under Ising model. $\alpha$ level is pre-set and dashed black line is TPR = 1.}
  \label{fig: EXP3 nonlinear fdr}
\end{figure}

The post-correction result is shown in Figure \ref{fig: EXP3 nonlinear fdr}. We note that when the graph estimate fundamentally misses true edges at $n=100$, the post-correction fails controlling the false discoveries. In comparison, we find in figure \ref{fig: EXP3 laplace fdr} and figure \ref{fig: EXP3 lognormal fdr} where linearity is satisfied, FDR is well controlled even when true edges are missed in the graph estimate. In contrast, we see when the true graph is contained in the graph estimate (the last column), ACV successfully filters out false discoveries, and is consistent for band, ER and SF graphs for nonlinear Ising model at $n=1000$. Even when the true graph is roughly contained at $n=500$, we observe that all FDR is well controlled even when the linearity assumption fails. In short, when the graph estimate contains or roughly contains the true graph, ACV has better pre-screening with less computational cost compared to BIC and EBIC. Incorporated with the post-correction, ACV outperforms BIC and EBIC.

\subsection{Tuning $\alpha$ in practice}
\label{sec:appendix alpha}
In practice, true graph is unknown, and choosing a proper $\alpha$ is tricky. On one hand, a small $\alpha$ is too restrictive, especially when $\wG\not\supseteq\G$, the non super-set case as shown in Figure~\ref{fig: EXP3 tpr fdr}. On the other hand, a large $\alpha$ fails the purpose of controlling false discoveries, e.g., the $\alpha = 0.9$ for $n=1000$ in Figure \ref{fig: EXP3 nonlinear fdr}.

Intuitively, model selection is one step beyond predictive tasks; A model with good predictive performance may not necessarily be the true model, but a true model shall always expect good predictive performance. The same intuition is well demonstrated in  Appendix F in \citep{stephenson2020approximate}. In other words, a smaller $\alpha$ is preferable if it achieves performance similar to other larger $\alpha$. Hence, we can evaluate $\alpha$ by comparing the predictive loss of the estimated graph $\check{\G}^{\alpha}$ with those by other $\alpha$ choices. Given $\wG^{\lambda}$ and $\alpha\in(0,1)$, we proceed the post-correction to have the estimated graph $\check{G}^{\alpha}$, and further refit the estimated graph and obtain a relative error compared to the error by $\wG^{\lambda}$. We demonstrate this with $\wG^{\lambdaacv}$ for $p=100$ case (the estimated graph by ACV $K$-fold in Figure \ref{fig: EXP3 tpr fdr}) for varying $\alpha\in[0.1, 0.9]$. We include $n\in\{100, 500, 1000\}$ to correspond the three super set relations between $\wG^{\lambdaacv}$ and $\G$.

We observe when the graph estimate is not a super set of the true graph ($n=100$), the relative error is extremely small but there is no regular trending. This warns the quality of the graph estimate we used for the post-correction, because a graph containing all true edges shall not expect erratic predictive performance. In contrast, we see a regular decreasing pattern in the other two cases when $\wG\mathrel{\substack{
  \textstyle\supset\\[-0.2ex]\textstyle\sim}}\G$ and $\wG\supseteq\G$ ($n=500,1000$). When the estimated graph is roughly a super set (the second column), the gap between small and large $\alpha$ remains even with $n$ increasing. This indicates that relaxing false discoveries constantly brings performance improvement, so we shall use a less restrictive $\alpha$ if we aim for predictive tasks. The observation aligns with the second column in Figure \ref{fig: EXP3 tpr fdr}, where we see TPR increases (but does not reach to 1) with increasing $\alpha$. In comparison, when the estimate graph is a super set of the true graph (the last column), the gap diminishes. This signals a only a marginal improvement from false discovery relaxation. Hence, we can expect a smaller $\alpha$ for roughly similar performance but better false discovery control.

  We note tuning $\alpha$ has two advantages over $\lambda$. First, $\alpha$ directly encodes a target error rate (e.g. an FDR level), whereas $\lambda$ is a scale parameter on the regularization path and does not map transparently to a desired error rate. Second, $\alpha$ lies in a fixed and interpretable range $(0,1)$, independent of the data scale or model dimension, while the appropriate range of $\lambda$ is problem‑dependent and typically chosen heuristically. Given $\alpha$, our procedure automatically selects $\lambda$ via ACV and then applies the $\alpha$‑level partial‑correlation test in the post‑correction step. %

\begin{table}[!ht]
  \caption{Comparison via F1 scores between structure learning after post-correction.}
  \label{tab:f1 by alpha}
  \begin{center}
    \begin{small}
      \begin{sc}
        \begin{tabular}{lccccccc}
          \toprule
          $\alpha$ & ACV $K$-fold & BIC & EBIC & CV $K$-fold & Lockhart's covariance test & STABILITY & TIGER \\
          \midrule
          0.1 & 0.977 & 0.977 & 0.915 & 0.977 & 0.977 & 0.013 & 0.977 \\
          0.3 & 0.990 & 0.987 & 0.919 & 0.990 & 0.987 & 0.013 & 0.990 \\
          0.5 & 0.997 & 0.990 & 0.923 & 0.997 & 0.990 & 0.013 & 0.997 \\
          0.7 & 0.994 & 0.997 & 0.923 & 0.994 & 0.990 & 0.013 & 0.994 \\
          0.9 & 0.990 & 0.993 & 0.923 & 0.990 & 0.990 & 0.013 & 0.990 \\
          \bottomrule
        \end{tabular}
      \end{sc}
    \end{small}
  \end{center}
  \vskip -0.1in
\end{table}
\subsection{Comparison with existing structure learning methods}
\label{sec:appendix other methods}
We add stability selection \citep[][]{liu2010stability3}, TIGER \citep[][]{liu2017tiger} and Lockhart’s covariance test~\citep[][]{lockhart2014significance} under the same pre-screening setup in section \ref{sec: knn pre-screen setup} to focus on the KNN graph with Gaussian data.  Our implementation is based on huge and flare R packages \citep[][]{flare,huge}. We also add F1 score as an aggregate metric for more straightforward comparison. The pre-screening result is shown in Figure \ref{fig: EXP2 knn compare}. We observe while the stability selection takes the least computation, it barely makes any meaning edge detections. Further we notice that EBIC outperforms ACV $K$-fold via F1 score when $p=500$, but it requires much more data than ACV $K$-fold to reach TPR = 1. When $n < p$, ACV $K$-fold provides the best edge detections with the least computation, signaling its outstanding pre-screening ability in the high-dimensional setting.

To better compare their overall performance after the post-correction, we focus on $p=100$ and $n=500$ under F1 score, and the result is shown in Table \ref{tab:f1 by alpha}. We first observe that with decreasing FDR control (increasing $\alpha$), the F1 score of the stability method remains the same. This signals its problem in pre-screening. We also notice that F1 scores of EBIC and Lockhart's covariance test are gradually increasing with $\alpha$, marking an effort in controlling false discoveries. In contrast, other methods all have F1 scores increasing and then decreasing. This implies that at a certain $\alpha$ level, all true positives haven been included, and further increasing $\alpha$ brings false discoveries and leads to lower F1 scores. Another way to interpret this is to only ACV $K$-fold, 3-fold CV and TIGER have reached the highest F1 scores with the smallest $\alpha$, and among those methods, ACV $K$-fold is the most computationally efficient one.

\end{document}